\documentclass[12pt,letterpaper]{article}
\usepackage[a4paper, total={7in, 10in}]{geometry}

\usepackage{graphicx}
\usepackage{helvet}
\usepackage{authblk}
\usepackage{hyperref}
\usepackage{amsmath} 
\usepackage{amssymb} 
\usepackage{orcidlink} 
\usepackage[super,comma,sort&compress]  
   {natbib}
\usepackage[right]{lineno} \linenumbers
\usepackage[title]{appendix}%

\usepackage[utf8]{inputenc} % allow utf-8 input
\usepackage[T1]{fontenc}    % use 8-bit T1 fonts
\usepackage{url}            % simple URL typesetting
\usepackage{booktabs}       % professional-quality tables
\usepackage{amsfonts}       % blackboard math symbols
\usepackage{nicefrac}       % compact symbols for 1/2, etc.
\usepackage{microtype}      % microtypography
\usepackage{xcolor}         % colors
\usepackage{multirow}
\usepackage{threeparttable}
\usepackage{amsmath}
\usepackage{subcaption}
\usepackage{makecell}
\usepackage{placeins}
\usepackage[utf8]{inputenc} % 支持 UTF-8 编码
\usepackage{booktabs}      % 用于生成专业的三线表
\usepackage{siunitx}       % 用于按小数点对齐数字
\usepackage{caption}       % 用于表格标题
\usepackage{tabularx}
\usepackage{listings}    % 引入 listings 宏包
\usepackage{pifont}
\usepackage{array}
\usepackage{makecell}
\theadset{\bfseries}

\makeatletter
\renewcommand{\maketitle}{\bgroup\setlength{\parindent}{0pt}
\begin{flushleft}
  \textbf{\@title}
  
  \@author
\end{flushleft}\egroup}
\makeatother

\title{ImputeECG: Deep Learning Reconstruction of Complete 12-Lead Electrocardiograms from Incomplete Recordings for Cardiac Assessment}
\date{}

\author[1,2,5]{Xiaocheng Fang}
\author[1,4,5]{Haoyu Wang}
\author[4]{Jieyi Cai}
\author[8]{Qinghao Zhao}
\author[1,5]{Jun Li}
\author[1,3]{Shanwei Zhang}
\author[1,2,5]{Guangkun Nie}
\author[1,5]{Yujie Xiao}
\author[1,5]{Shun Huang}
\author[1,2,5]{Jiarui Jin}
\author[9]{Hongmin Liu}
\author[9]{Guodong Wang}
\author[9]{Shuohua Chen}
\author[9]{Liming Lin}
\author[9]{Shouling Wu}
\author[2,*]{Hongyan Li}
\author[1,5,6,7*]{Shenda Hong}

\affil[1]{National Institute of Health Data Science, Peking University, Beijing, China}
\affil[2]{School of Intelligence Science and Technology, Peking University, Beijing, China}
\affil[3]{Department of Computer Science, Tianjin University of Technology, Tianjin, China}
\affil[4]{University of the Chinese Academy of Sciences, Beijing, China}
\affil[5]{Institute of Medical Technology, Peking University Health Science Center, Beijing, China}
\affil[6]{State Key Laboratory of Vascular Homeostasis and Remodeling, NHC Key Laboratory of Cardiovascular Molecular Biology and Regulatory Peptides, Peking University, Beijing, China} 
\affil[7]{Institute for Artificial Intelligence, Peking University, Beijing, China}
\affil[8]{Department of Cardiology, Peking University People’s Hospital, Beijing, China}
\affil[9]{Department of Cardiology, Kailuan General Hospital, Tangshan, China}

\affil[*]{Correspondence: hongshenda@pku.edu.cn}

\begin{document}

\maketitle

\section*{Abstract}
Complete digital 12-lead electrocardiograms (ECGs) are essential for AI-enabled cardiovascular assessment, yet many clinical ECG records, particularly those digitized from ECG images, remain incomplete because of short display formats, incomplete waveform digitization, lead loss, or signal corruption. We developed ImputeECG, a mask-conditioned one-dimensional Transformer autoencoder that completes 12-lead, 10-s ECGs while retaining all observed samples. The model was trained on PTB-XL and evaluated on PTB-XL and CPSC2018 under simulated incomplete settings, with additional real-world validation in a 43{,}633-record Kailuan clinical cohort after ECG image digitization. Metrics were computed over originally missing regions, with analyses of morphology and downstream diagnostic utility. On PTB-XL, ImputeECG reduced missing-region MAE by 41.7--51.0\% and MSE by 54.0--63.7\% versus the strongest baseline, with lower errors in R-peak timing, RR interval, QRS duration, QT interval, and P-wave, QRS-complex, and T-wave reconstruction. On CPSC2018, ImputeECG reduced MAE by 49.7--51.9\%, supporting external generalization. In downstream multi-label classification, ImputeECG restored performance to 92.28\% AUROC and 33.88\% AUPRC in the most incomplete PTB-XL setting, approaching complete-ECG performance. On CPSC2018, completed ECGs achieved 94.75--95.89\% AUROC and 78.83--81.86\% AUPRC across settings. In Kailuan, ECG completion improved zero-shot sex prediction AUROC from 82.6\% to 85.8\% and reduced age prediction MAE from 10.72 to 9.87 years after image-based ECG digitization. These findings support ECG completion as a practical strategy for converting incomplete ECG records into AI-ready 12-lead, 10-s digital signals and extending the usable scope of ECG archives for digital cardiac assessment.

% 完整的12导联数字心电图（ECG）对于人工智能辅助的心血管评估至关重要，然而，由于显示格式短、数字化不完整、导联丢失或信号损坏等原因，许多临床心电图记录仍仅部分可见。我们开发了ImputeECG，这是一种基于掩码条件的一维Transformer自编码器，能够完整呈现12导联、10秒的心电图，同时保留所有可观测的样本。该模型在PTB-XL数据集上进行训练，并在模拟不完整设置下于PTB-XL和CPSC2018数据集上进行评估。此外，在心电图图像数字化后，我们还在一个包含43,633条记录的开滦临床队列中进行了真实世界验证。我们计算了原始缺失区域的指标，并分析了其形态和下游诊断效用。在 PTB-XL 数据集上，与最强的基线相比，ImputeECG 将缺失区域的平均绝对误差 (MAE) 降低了 41.7% 至 51.0%，平均误差 (MSE) 降低了 54.0% 至 63.7%，并且在 R 波峰值时间、RR 间期、QRS 波时限、QT 间期以及 P 波、QRS 波群和 T 波的重建方面均有更低的误差。在 CPSC2018 数据集上，ImputeECG 将 MAE 降低了 49.7% 至 51.9%，支持外部泛化。在下游多标签分类中，即使在数据最不完整的 PTB-XL 设置下，ImputeECG 也能将性能恢复至 92.28% 的 AUROC 和 33.88% 的 AUPRC，接近完整心电图的性能。在 CPSC2018 数据集上，完整心电图在所有设置下均实现了 94.75% 至 95.89% 的 AUROC 和 78.83% 至 81.86% 的 AUPRC。在开滦，基于图像的心电图数字化后，心电图补全将零样本性别预测的AUROC从82.6%提高到85.8%，并将年龄预测的平均绝对误差（MAE）从10.72年降低到9.87年。这些发现支持将心电图补全作为一种实用策略，用于将不完整的心电图记录转换为可用于人工智能的12导联、10秒数字信号，并扩展心电图档案在数字心脏评估中的可用范围。

\section*{Keywords}
12-lead Electrocardiography, Incomplete ECG Records, ECG Signal Completion, Mask-conditioned Transformer, AI-enabled Cardiac Assessment

\section*{Introduction}
The standard 12-lead electrocardiogram (ECG) remains one of the most widely used diagnostic tests in cardiovascular medicine. A complete 10-second digital ECG provides synchronized temporal and spatial information across limb and precordial leads, supporting rhythm interpretation, conduction assessment, ischemia detection, chamber abnormality evaluation, and longitudinal disease monitoring~\cite{kligfield2007recommendations,hannun2019cardiologist,ribeiro2020automatic,jin2025self}. With the rapid development of artificial intelligence (AI) for ECG analysis, complete digital waveforms have also become a critical substrate for automated diagnosis, risk prediction, disease screening, and large-scale phenotyping~\cite{attia2019artificial,raghunath2020prediction,wu2022fully,jin2026ecg}. As AI-enabled ECG interpretation moves from retrospective research toward real-world clinical deployment, the availability of complete, standardized, multi-lead ECG signals has become increasingly important.

% 标准12导联心电图（ECG）依然是心血管医学中最广泛应用的诊断检查之一。完整的10秒数字心电图记录提供了涵盖肢体导联和胸前导联的同步时空信息，支持心律判读、传导功能评估、缺血检测、心腔异常评估以及疾病的纵向监测~\cite{kligfield2007recommendations,hannun2019cardiologist,ribeiro2020automatic,jin2025self}。随着人工智能（AI）心电分析技术的飞速发展，完整的数字波形已成为实现自动诊断、风险预测、疾病筛查及大规模表型分析的关键基础~\cite{attia2019artificial,raghunath2020prediction,wu2022fully}。在AI辅助心电判读从回顾性研究向真实世界临床应用转型的过程中，获取完整、标准化且包含多导联信息的ECG信号变得愈发重要。

In routine practice, however, a substantial proportion of ECG data remains incomplete. Many hospitals and health systems continue to archive ECGs as paper printouts, scanned images, or PDF reports, particularly in retrospective cohorts and settings where native digital waveform storage is limited~\cite{stenhede2026digitizing,wu2022fully,fortune2022digitizing}. Although ECG digitization algorithms can convert visible waveform traces into digital time-series signals, they are intrinsically constrained by the waveform segments displayed in the original report. Common standard ECG printout formats often show short segments of each lead, such as 2.5-second recordings for the 12 leads accompanied by a 10-second rhythm strip, whereas complete digital ECG datasets typically preserve synchronized 12-lead, 10-second waveforms~\cite{gliner2020automatic,wagner2020ptb}. As a result, large portions of the underlying multi-lead signal may remain unavailable after digitization. ECG incompleteness may also arise when leads or waveform regions are blank, degraded, or unrecoverable during image-based digitization workflows~\cite{lence2025ecgrecover}. These missing waveforms restrict the reuse of historical ECG archives, reduce the compatibility of incomplete digitized ECGs with modern AI models, and create a practical barrier to constructing large-scale digital cardiac datasets~\cite{siontis2021artificial,ribeiro2020automatic}.

% 然而，在常规临床实践中，相当一部分心电图（ECG）数据仍处于不完整或仅部分可观测的状态。许多医院和医疗系统仍将心电图存档为纸质打印件、扫描图像或PDF报告，尤其是在回顾性队列研究及原生数字波形存储能力有限的场景中~\cite{stenhede2026digitizing,wu2022fully}。尽管心电图数字化算法能将可见的波形轨迹转换为数字时间序列信号，但其效果本质上受限于原始报告中展示的波形片段。常见的标准心电图打印格式通常仅显示各导联的短片段（例如12导联各2.5秒的记录外加一条10秒的节律条），而完整的心电图数字数据集通常保留同步的12导联、10秒波形数据~\cite{gliner2020automatic,wagner2020ptb}。因此，数字化后，很大一部分原始多导联信号可能仍无法获取。此外，在基于图像的数字化流程中，若某些导联或波形区域出现空白、质量退化或无法恢复的情况，也会导致数据仅呈部分可观测状态~\cite{lence2025ecgrecover}。这些缺失的波形限制了历史心电图档案的再利用，降低了部分数字化心电图与现代人工智能模型的兼容性，并构成了构建大规模数字心脏数据集的实际障碍~\cite{siontis2021artificial,ribeiro2020automatic}。

\begin{figure*}[t]
\centerline{\includegraphics[width=0.95\textwidth]{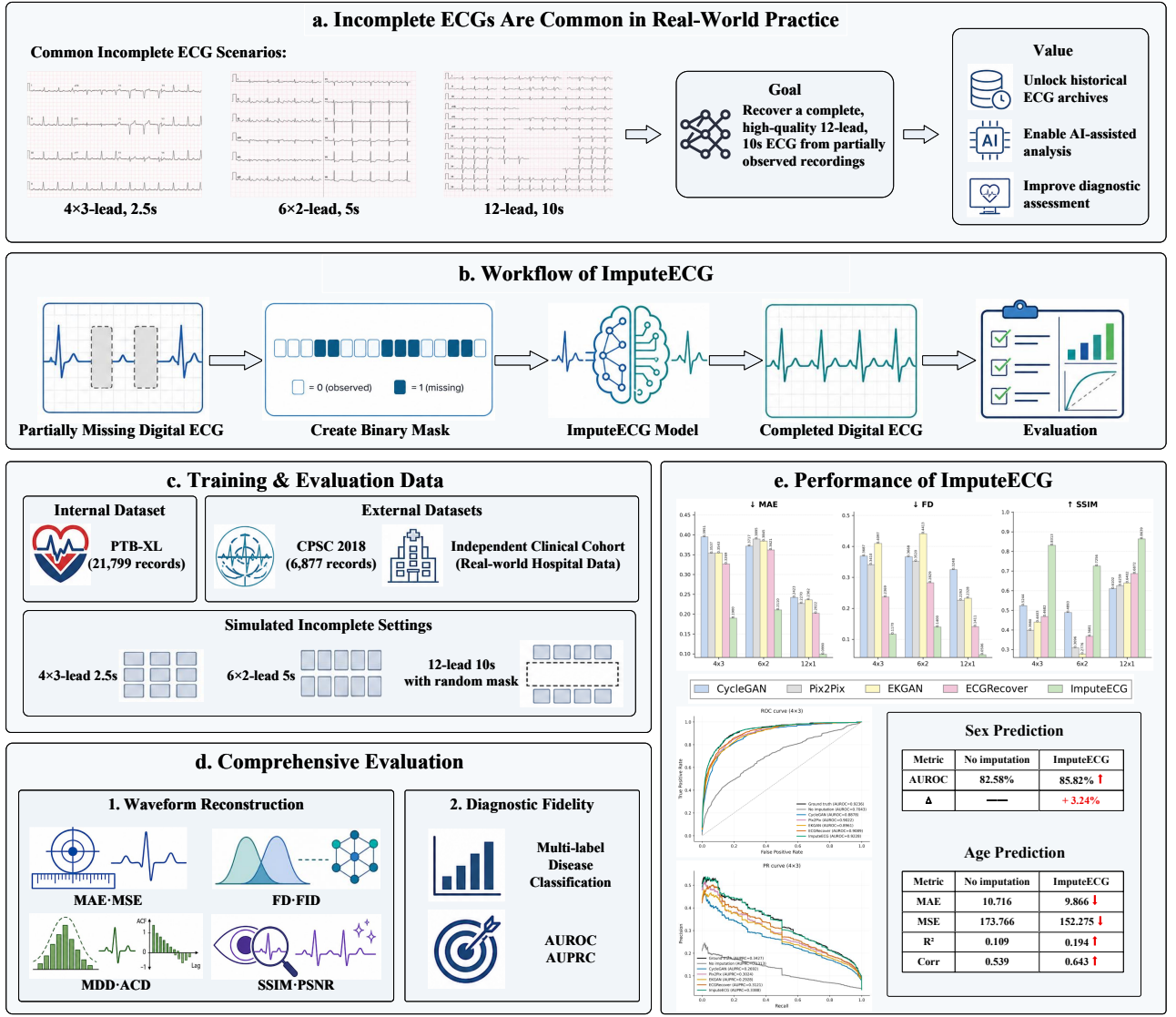}}
    \caption{
    \textbf{Overview of ImputeECG.}
    \textbf{a}, Real-world incomplete ECG scenarios and the goal of restoring complete 12-lead, 10-second ECGs.
    \textbf{b}, Workflow of mask-guided ECG completion.
    \textbf{c}, Datasets and simulated missingness settings.
    \textbf{d}, Evaluation framework for reconstruction quality and diagnostic fidelity.
    \textbf{e}, Representative performance comparison between ImputeECG and baseline methods.
    }
\label{fig:overview}
\end{figure*}

Existing computational approaches address related problems from several perspectives. ECG digitization algorithms extract visible waveforms from paper or image-based ECGs~\cite{li2020deep,shivashankara2024ecg}. Generic signal completion and image-to-image translation models~\cite{fang2025ppgflowecg,fang2026ecgflowcmr}, including generative adversarial networks and encoder–decoder architectures, have also been adapted to reconstruct missing signal regions~\cite{zhu2017unpaired,isola2017image}. Lead synthesis methods estimate unobserved leads from available leads by exploiting physiological relationships among ECG leads~\cite{golany202112,joo2023twelve,chen2024multi,lee2025parameter,lence2025ecgrecover}. Despite these advances, a clinically useful ECG completion framework must address several requirements simultaneously. ECGRecover~\cite{lence2025ecgrecover} has provided an important step toward formalizing ECG completion by considering both segment recovery and lead reconstruction, using realistic masking scenarios and evaluating waveform-level fidelity and ECG landmark preservation. However, clinical ECG archives introduce a broader completion problem in which incomplete recordings may arise from short-display formats, image digitization failures, lead loss or local signal corruption. In this setting, a reconstruction method should explicitly condition on the observed-sample mask, model both cross-lead and long-range temporal dependencies, preserve the originally observed waveform segments exactly, and generate completed signals that remain useful for downstream clinical interpretation. A central remaining need is therefore a unified ECG reconstruction approach evaluated across waveform-level fidelity, morphology preservation and task-level clinical utility under both internal and external validation settings.

% 现有的计算方法从多个角度解决了相关问题。心电图数字化算法从纸质或图像心电图中提取可见波形~\cite{li2020deep,shivashankara2024ecg}。通用信号补全和图像到图像的转换模型，包括生成对抗网络和编码器-解码器架构，也被用于重建缺失的信号区域~\cite{zhu2017unpaired,isola2017image}。导联合成方法利用心电图导联之间的生理关系，根据现有导联估计未观测到的导联~\cite{joo2023twelve,lence2025ecgrecover}。尽管取得了这些进展，但一个临床实用的心电图补全框架必须同时满足多个要求。 ECGRecover~\cite{lence2025ecgrecover} 通过考虑节段恢复和导联重建，并采用真实的掩蔽场景，评估波形级保真度和心电图标志点保留情况，为心电图补全的规范化迈出了重要一步。然而，临床心电图存档引入了更广泛的补全问题，其中不完整的记录可能由短显示格式、图像数字化失败、导联丢失或局部信号损坏等原因造成。在这种情况下，重建方法应明确地以观测样本掩蔽为条件，对跨导联和长程时间依赖性进行建模，精确保留原始观测到的波形节段，并生成可用于后续临床解读的完整信号。因此，目前亟需一种统一的心电图重建方法，并在内部和外部验证设置下，从波形级保真度、形态保留和任务级临床实用性三个方面进行评估。

To this end, we present ImputeECG, a mask-guided deep learning framework for reconstructing complete 12-lead, 10-second ECGs from incomplete recordings. ImputeECG uses a one-dimensional Vision Transformer encoder–decoder architecture that takes the observed ECG signal and its corresponding missingness mask as input. By jointly modeling temporal context and inter-lead dependencies, the model learns to infer missing waveform regions while retaining the original observed segments. This design aligns with real-world ECG completion scenarios, including short multi-lead segments, half-lead split recordings, and localized missing intervals within individual leads.

% 为此，我们提出了 ImputeECG，一个基于掩码引导的深度学习框架，用于从部分观测到的心电图记录中重建完整的 12 导联 10 秒心电图。ImputeECG 采用一维 Vision Transformer 编码器-解码器架构，以观测到的心电信号及其对应的缺失掩码作为输入。通过联合建模时间上下文和导联间依赖关系，该模型能够学习推断缺失的波形区域，同时保留原始观测片段。这种设计符合实际的心电图补全场景，包括短的多导联片段、半导联分割记录以及单个导联内的局部缺失间期。

We evaluated ImputeECG using simulated incomplete ECG settings derived from PTB-XL~\cite{wagner2020ptb} and external testing on CPSC2018~\cite{liu2018open}. The missingness settings were designed to reflect clinically relevant acquisition and digitization patterns, including \(4\times3\)-lead 2.5-second display layouts, \(6\times2\)-lead 5-second display layouts, and 12-lead 10-second recordings with local temporal signal gaps. ImputeECG was compared with representative reconstruction baselines, including CycleGAN~\cite{zhu2017unpaired}, Pix2Pix~\cite{isola2017image}, EKGAN~\cite{joo2023twelve}, and ECGRecover~\cite{lence2025ecgrecover}. Reconstruction quality was assessed using waveform-level error metrics, and morphology preservation was further evaluated on PTB-XL to determine whether completed ECGs retained clinically relevant waveform characteristics. Beyond signal fidelity, we evaluated whether completed ECGs preserved clinically relevant information for downstream prediction tasks. On PTB-XL, we assessed multi-label diagnostic subclass classification; on CPSC2018, we assessed external multi-label rhythm and conduction classification. For both datasets, classifiers were applied to complete reference ECGs, masked ECGs, and imputed ECGs generated by each reconstruction method, allowing direct comparison of downstream AUROC and AUPRC across input conditions. We further evaluated ImputeECG in the independent Kailuan clinical cohort, a real-world collection of digitized ECGs containing short-display formats and occasional corrupted or unrecovered waveform segments. In this cohort, pretrained Net1D models were applied directly to digitized ECGs before and after completion for sex prediction and age prediction, thereby assessing whether ECG completion improved the downstream usability of incomplete digitized clinical ECG archives.

% 我们使用源自 PTB-XL~\cite{wagner2020ptb} 的模拟不完整心电图设置以及在 CPSC2018~\cite{liu2018open} 上的外部测试对 ImputeECG 进行了评估。缺失值设置旨在反映临床相关的采集和数字化模式，包括 4×3 导联 2.5 秒显示布局、6×2 导联 5 秒显示布局以及具有局部时间信号间隙的 12 导联 10 秒记录。我们将 ImputeECG 与具有代表性的重建基线进行了比较，包括 CycleGAN~\cite{zhu2017unpaired}、Pix2Pix~\cite{isola2017image}、EKGAN~\cite{joo2023twelve} 和 ECGRecover~\cite{lence2025ecgrecover}。我们使用波形级误差指标评估了重建质量，并在PTB-XL数据集上进一步评估了形态保留情况，以确定重建后的心电图是否保留了临床相关的波形特征。除了信号保真度之外，我们还评估了重建后的心电图是否保留了用于下游预测任务的临床相关信息。在PTB-XL数据集上，我们评估了多标签诊断亚类分类；在CPSC2018数据集上，我们评估了外部多标签节律和传导分类。对于这两个数据集，我们将分类器应用于由每种重建方法生成的完整参考心电图、掩蔽心电图和插补心电图，从而可以直接比较不同输入条件下的下游AUROC和AUPRC。我们还在独立的开滦临床队列中评估了ImputeECG方法，该队列是一个包含短显示格式和偶尔损坏或未恢复的波形片段的真实世界数字化心电图集合。在本研究中，预训练的 Net1D 模型被直接应用于数字化心电图，用于性别预测和年龄预测，从而评估心电图的完成是否提高了不完整的数字化临床心电图档案的下游可用性。

By formulating incomplete ECG restoration as a clinically grounded digital medicine problem, this study provides a framework for transforming incomplete ECG records into complete digital signals suitable for AI-assisted analysis. The ability to recover complete 12-lead ECGs from incomplete recordings may expand the value of historical ECG archives, improve the usability of digitized paper ECGs, and support broader development of scalable cardiovascular AI systems across heterogeneous clinical environments~\cite{nie2025anyppg}.
% 本研究将不完整心电图恢复问题转化为一个具有临床意义的数字医学问题，并以此为基础构建了一个框架，用于将部分可用的心电图记录转换为适用于人工智能辅助分析的完整数字信号。从不完整记录中恢复完整12导联心电图的能力，有望提升历史心电图档案的价值，提高数字化纸质心电图的可用性，并支持在各种不同的临床环境中更广泛地开发可扩展的心血管人工智能系统。

\section*{Results}

% \subsection*{Overview of the ImputeECG Framework and Evaluation Design}

% We developed ImputeECG to reconstruct complete 12-lead, 10-second electrocardiograms (ECGs) from partially observed recordings. The model was evaluated under three clinically motivated incomplete-recording settings: a 2.5-s 4x3 layout, in which each 2.5-s window retained only three consecutive leads; a 5-s 6x2 layout, in which the first and second halves of the recording retained complementary six-lead subsets; and a 10-s 12x1 layout, in which each lead contained a continuous masked segment. These settings were designed to emulate incomplete ECGs arising from short display windows, split lead acquisition, and localized waveform loss.

% PTB-XL was used for model development and internal testing, with 52,254 recordings for training, 6,549 for validation, and 6,594 for testing. CPSC2018 was used as an external reconstruction dataset with 6,877 ECGs, and an additional CPSC2018 classification experiment was conducted using a 7:1:2 train-validation-test split. ImputeECG was compared with representative reconstruction baselines, including CycleGAN~\cite{zhu2017unpaired}, Pix2Pix~\cite{isola2017image}, EKGAN~\cite{joo2023twelve}, and ECGRecover~\cite{lence2025ecgrecover}. Performance was assessed at two levels: waveform fidelity using mean absolute error (MAE) and root mean squared error (RMSE), and downstream clinical utility using AUROC and AUPRC for ECG classification.

\begin{figure*}[t]
\centerline{\includegraphics[width=1\textwidth]{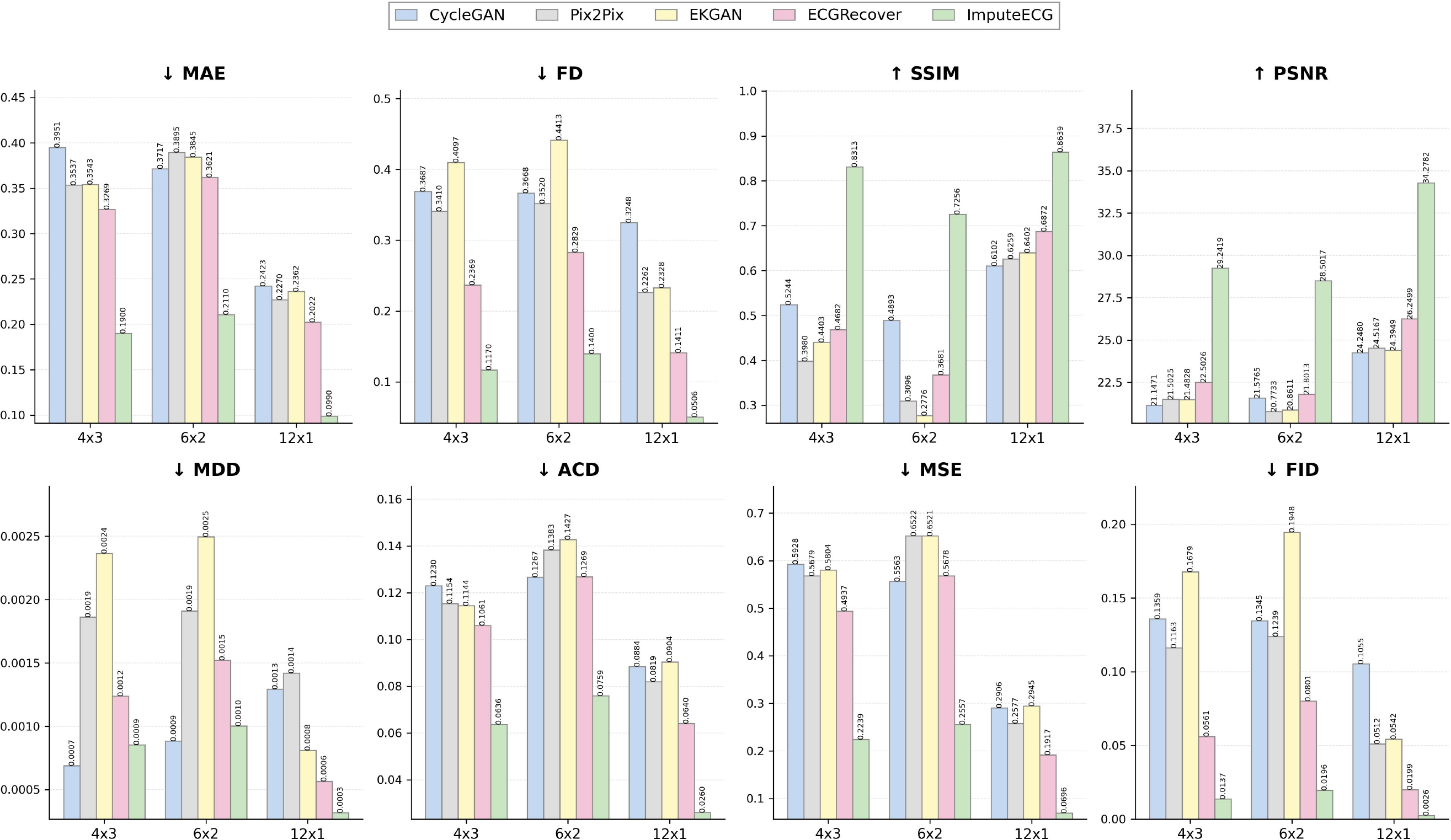}}
\caption{
    \textbf{Reconstruction performance on the PTB-XL internal test set.}
    Bar plots compare ImputeECG with other baselines under three incomplete ECG settings.
    Metrics were computed over originally missing regions.
    Lower values indicate better performance for MAE, FD, MDD, ACD, MSE, and FID; higher values indicate better performance for SSIM and PSNR.
    ImputeECG showed the strongest overall reconstruction performance across settings.
    }
\label{fig:ptbxl_reconstruction}
\end{figure*}

\subsection*{ImputeECG Improved Waveform Reconstruction on PTB-XL}

On the internal PTB-XL test set, the proposed ImputeECG outperformed the other baselines by a substantial margin across most reconstruction metrics (Fig.~\ref{fig:ptbxl_reconstruction}). All metrics were computed only over the originally missing regions. For point-wise reconstruction accuracy, ImputeECG achieved the lowest MAE values of 0.190, 0.211, and 0.099 in the \(4 \times 3\), \(6 \times 2\), and \(12 \times 1\) settings, respectively, corresponding to relative reductions of 41.9\%, 41.7\%, and 51.0\% compared with the strongest baseline in each setting. MSE showed a consistent improvement pattern, with relative reductions of 54.6\%, 54.0\%, and 63.7\% across the three settings.
% 在内部测试集 PTB-XL 上，所提出的 ImputeECG 在大多数重建指标上均显著优于其他基线方法（图~\ref{fig:ptbxl_reconstruction}）。所有指标均仅针对原始缺失区域进行计算。对于逐点重建精度，ImputeECG 在 \(4 \times 3\)、\(6 \times 2\) 和 \(12 \times 1\) 设置下分别实现了最低的 MAE 值，分别为 0.190、0.211 和 0.099，与各设置下最强的基线相比，相对降低了 41.9%、41.7% 和 51.0%。MSE 也呈现出一致的改进趋势，在三种设置下分别降低了 54.6%、54.0% 和 63.7%。

Beyond point-wise error, ImputeECG also showed consistent advantages in waveform similarity, temporal consistency, structural preservation, and signal quality. Across the \(4 \times 3\), \(6 \times 2\), and \(12 \times 1\) settings, FD was reduced by 50.6\%, 50.5\%, and 64.1\%, ACD by 40.1\%, 40.1\%, and 59.4\%, and FID by 75.6\%, 75.5\%, and 86.9\% relative to the strongest baseline. Metrics where higher values indicate better reconstruction also improved consistently: SSIM increased by 0.307, 0.236, and 0.177, and PSNR increased by 6.74, 6.70, and 8.03~dB across the three settings.
% 除了逐点误差外，ImputeECG 在波形相似性、时间一致性、结构保留和信号质量方面也表现出持续的优势。在 4×3、6×2 和 12×1 三种设置下，与最强基线相比，FD 分别降低了 50.6%、50.5% 和 64.1%，ACD 分别降低了 40.1%、40.1% 和 59.4%，FID 分别降低了 75.6%、75.5% 和 86.9%。其他指标（数值越高表示重建效果越好）也持续改善：SSIM 分别提高了 0.307、0.236 和 0.177，PSNR 分别提高了 6.74、6.70 和 8.03 dB。

MDD showed a more setting-dependent pattern. ImputeECG achieved the lowest MDD in the \(12 \times 1\) setting, with a value of 0.0003 compared with 0.0006 for the strongest baseline, and remained close to the best-performing baseline in the \(4 \times 3\) and \(6 \times 2\) settings. Overall, these results indicate that ImputeECG provided robust reconstruction gains across point-wise error, temporal dependence, structural similarity, signal-to-noise quality, and Fr\'echet-based distributional distance, with smaller setting-dependent differences in marginal amplitude distribution.
% MDD 呈现出更明显的设置依赖性。在 \(12 \times 1\) 设置下，ImputeECG 的 MDD 值最低，为 0.0003，而最强基线的 MDD 值为 0.0006；在 \(4 \times 3\) 和 \(6 \times 2\) 设置下，ImputeECG 的 MDD 值也接近最佳基线。总体而言，这些结果表明，ImputeECG 在逐点误差、时间依赖性、结构相似性、信噪比以及基于 Fréchet 分布距离等方面均提供了稳健的重建增益，且边缘振幅分布的设置依赖性差异较小。

\begin{figure*}[t]
\centerline{\includegraphics[width=1\textwidth]{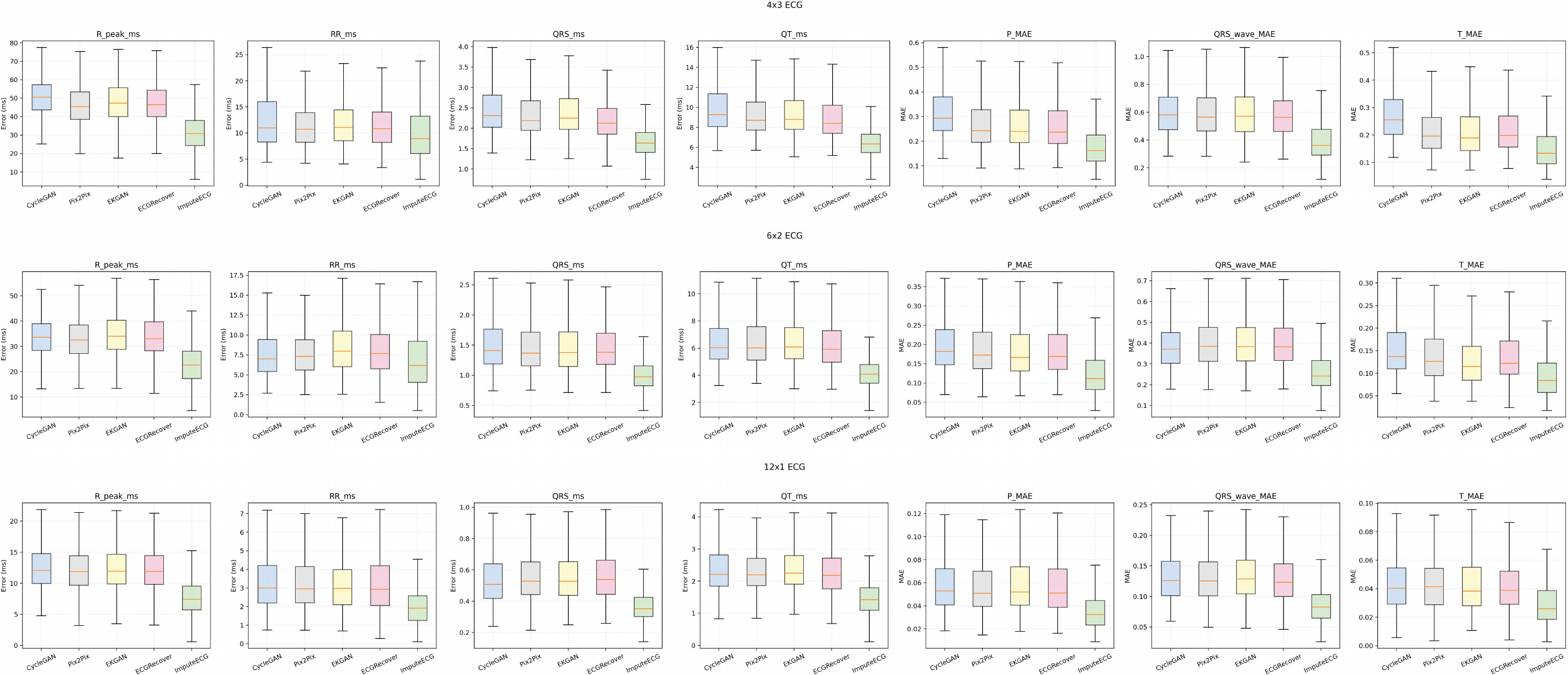}}
\caption{ \textbf{Morphology preservation on PTB-XL.} Box plots compare reconstruction errors in ECG timing features and waveform regions across three incomplete ECG settings. Lower values indicate better preservation of ECG morphology. }
\label{fig:ptbxl_morphology}
\end{figure*}

Morphology-based evaluation further supported the reconstruction results (Fig.~\ref{fig:ptbxl_morphology}). Across the three incomplete ECG settings, ImputeECG showed lower median errors for clinically relevant timing features, including R-peak timing, RR interval, QRS duration, and QT interval. The same trend was observed for waveform-region errors, where ImputeECG reduced MAE in P-wave, QRS-complex, and T-wave segments. These findings indicate that ImputeECG preserved both global waveform fidelity and morphology-related ECG features on PTB-XL, supporting its ability to reconstruct missing signal regions while maintaining clinically meaningful waveform structure.
% 基于形态学的评估进一步支持了重建结果（图~\ref{fig:ptbxl_morphology}）。在三种不完整心电图设置中，ImputeECG 在临床相关的时序特征（包括 R 波峰值时间、RR 间期、QRS 波时限和 QT 间期）方面均显示出较低的中位误差。波形区域误差也观察到相同的趋势，ImputeECG 降低了 P 波、QRS 波群和 T 波段的平均绝对误差 (MAE)。这些发现表明，ImputeECG 在 PTB-XL 上既保留了整体波形保真度，又保留了与形态学相关的心电图特征，这支持了其在重建缺失信号区域的同时保持临床意义波形结构的能力。

\subsection*{Completed ECGs Restored Downstream Diagnostic Performance on PTB-XL}

We next evaluated whether reconstructed ECGs preserved diagnostic information in downstream multi-label classification. Complete ground-truth ECGs provided the reference performance on PTB-XL, with an AUROC of 92.36\% (95\% CI, 91.31--93.31) and an AUPRC of 34.27\% (95\% CI, 34.15--38.76) (Table~\ref{tab:ptbxl_downstream_ci}). Direct classification on masked ECGs showed substantial degradation, especially in the \(4 \times 3\) setting, where AUROC decreased to 70.43\% (95\% CI, 68.37--72.87) and AUPRC decreased to 13.13\% (95\% CI, 12.65--16.08). This performance loss reflects the practical risk of directly analyzing incomplete or image-digitized short-display ECGs, supporting the need to complete such records into standardized 12-lead, 10-second ECG signals before downstream AI assessment.
% 接下来，我们评估了重建的心电图是否能在下游多标签分类中保留诊断信息。完整的真实心电图在PTB-XL数据集上提供了参考性能，其AUROC为92.36%（95% CI，91.31-93.31），AUPRC为34.27%（95% CI，34.15-38.76）（表）。直接对掩蔽心电图进行分类显示出显著的性能下降，尤其是在4×3设置下，AUROC下降至70.43%（95% CI，68.37-72.87），AUPRC下降至13.13%（95% CI，12.65-16.08）。这种性能损失反映了直接分析部分观察到的或图像数字化的短显示心电图的实际风险，也支持在下游 AI 评估之前将此类记录补充为标准化的 12 导联 10 秒心电图信号的必要性。

ImputeECG restored diagnostic performance most prominently in the \(4 \times 3\) setting. Classification using ImputeECG-completed ECGs achieved an AUROC of 92.28\% (95\% CI, 91.19--93.28) and an AUPRC of 33.88\% (95\% CI, 33.43--38.20), corresponding to absolute gains of 21.85 and 20.75 percentage points over masked inputs, respectively. These values approached the complete-ECG reference performance and exceeded the strongest reconstruction baseline, ECGRecover, which achieved 90.89\% AUROC and 31.21\% AUPRC.
% ImputeECG 在 \(4 \times 3\) 设置下显著提升了诊断性能。使用 ImputeECG 补全的心电图进行分类，其 AUROC 为 92.28%（95% CI，91.19--93.28），AUPRC 为 33.88%（95% CI，33.43--38.20），分别比掩蔽输入提高了 21.85 和 20.75 个百分点。这些数值接近完整心电图的参考性能，并超过了性能最佳的重建基线 ECGRecover（其 AUROC 为 90.89%，AUPRC 为 31.21%）。

In the \(6 \times 2\) setting, masked ECGs achieved an AUROC of 87.43\% (95\% CI, 85.99--88.53) and an AUPRC of 27.41\% (95\% CI, 26.97--31.47). ImputeECG increased performance to 92.25\% AUROC (95\% CI, 91.18--93.24) and 34.18\% AUPRC (95\% CI, 33.68--38.54), corresponding to absolute gains of 4.82 and 6.77 percentage points over masked inputs. It achieved the best reconstruction-based result in this setting and closely matched the complete-ECG reference.
% 在 \(6 \times 2\) 设置下，掩蔽心电图的 AUROC 为 87.43%（95% CI，85.99--88.53），AUPRC 为 27.41%（95% CI，26.97--31.47）。ImputeECG 将性能提升至 AUROC 92.25%（95% CI，91.18--93.24）和 AUPRC 34.18%（95% CI，33.68--38.54），与掩蔽输入相比，绝对增益分别为 4.82 和 6.77 个百分点。在此设置下，ImputeECG 取得了最佳的基于重建的结果，并且与完整心电图参考值高度吻合。

In the \(12 \times 1\) local-missingness setting, masked ECGs already retained relatively high diagnostic performance, with an AUROC of 91.88\% (95\% CI, 90.80--92.81) and an AUPRC of 33.27\% (95\% CI, 32.83--37.66). ImputeECG further increased AUROC to 92.39\% (95\% CI, 91.32--93.35), yielding the highest reconstruction-based AUROC and numerically exceeding the complete-ECG reference estimate. Its AUPRC reached 34.10\% (95\% CI, 34.04--38.60), close to the complete-ECG reference and comparable to the best reconstruction-based AUPRC. Overall, ImputeECG-completed ECGs preserved clinically relevant diagnostic information across missingness patterns, with the largest gains observed under the most severe incomplete-recording setting.
% 在局部缺失值比例为 12 × 1 的情况下，掩蔽心电图已保持了相对较高的诊断性能，AUROC 为 91.88%（95% CI，90.80-92.81），AUPRC 为 33.27%（95% CI，32.83-37.66）。ImputeECG 进一步将 AUROC 提高至 92.39%（95% CI，91.32-93.35），获得了最高的基于重建的 AUROC，并且在数值上超过了完整心电图的参考估计值。其 AUPRC 达到 34.10%（95% CI，34.04-38.60），接近完整心电图的参考值，并且与最佳的基于重建的 AUPRC 相当。总体而言，ImputeECG 完成的心电图保留了临床相关的诊断信息，即使在数据缺失的情况下也能保持信息完整，在最严重的局部观察设置下，增益最大。

\newcommand{\ci}[2]{#1 {\scriptsize [#2]}}
\newcommand{\cibest}[2]{\textbf{#1} {\scriptsize [#2]}}

\begin{table*}[t]
\centering
\caption{
\textbf{Downstream diagnostic performance on PTB-XL.}
AUROC and AUPRC are reported as percentages with 95\% bootstrap percentile confidence intervals.
Complete and masked ECGs are included as reference baselines.
Bold values indicate the best reconstruction-based result within each incomplete ECG setting.
}
\label{tab:ptbxl_downstream_ci}
\footnotesize
\setlength{\tabcolsep}{5pt}
\renewcommand{\arraystretch}{1.10}

\begin{tabular}{@{}lllcc@{}}
\toprule
\textbf{ECG setting} &
\textbf{Test input} &
\textbf{Method} &
\textbf{AUROC (\%, 95\% CI)} &
\textbf{AUPRC (\%, 95\% CI)} \\
\midrule

Complete ECG
& Ground truth
& --
& \ci{92.36}{91.31, 93.31}
& \ci{34.27}{34.15, 38.76} \\

\midrule
\multirow{6}{*}{\(4 \times 3\) ECG}
& Masked input
& No imputation
& \ci{70.43}{68.37, 72.87}
& \ci{13.13}{12.65, 16.08} \\
& \multirow{5}{*}{Reconstructed}
& CycleGAN~\cite{zhu2017unpaired}
& \ci{88.78}{87.55, 89.91}
& \ci{26.92}{26.78, 31.02} \\
& & Pix2Pix~\cite{isola2017image}
& \ci{90.22}{88.87, 91.24}
& \ci{30.24}{30.05, 34.49} \\
& & EKGAN~\cite{joo2023twelve}
& \ci{89.61}{88.29, 90.56}
& \ci{29.28}{29.14, 33.58} \\
& & ECGRecover~\cite{lence2025ecgrecover}
& \ci{90.89}{89.50, 91.81}
& \ci{31.21}{30.69, 35.59} \\
& & ImputeECG
& \cibest{92.28}{91.19, 93.28}
& \cibest{33.88}{33.43, 38.20} \\

\midrule
\multirow{6}{*}{\(6 \times 2\) ECG}
& Masked input
& No imputation
& \ci{87.43}{85.99, 88.53}
& \ci{27.41}{26.97, 31.47} \\
& \multirow{5}{*}{Reconstructed}
& CycleGAN~\cite{zhu2017unpaired}
& \ci{90.82}{89.58, 91.83}
& \ci{31.14}{30.88, 35.11} \\
& & Pix2Pix~\cite{isola2017image}
& \ci{91.46}{90.21, 92.36}
& \ci{32.97}{32.30, 37.28} \\
& & EKGAN~\cite{joo2023twelve}
& \ci{91.06}{89.71, 91.97}
& \ci{31.97}{31.54, 36.21} \\
& & ECGRecover~\cite{lence2025ecgrecover}
& \ci{91.57}{90.27, 92.49}
& \ci{33.17}{32.78, 37.63} \\
& & ImputeECG
& \cibest{92.25}{91.18, 93.24}
& \cibest{34.18}{33.68, 38.54} \\

\midrule
\multirow{6}{*}{\(12 \times 1\) ECG}
& Masked input
& No imputation
& \ci{91.88}{90.80, 92.81}
& \ci{33.27}{32.83, 37.66} \\
& \multirow{5}{*}{Reconstructed}
& CycleGAN~\cite{zhu2017unpaired}
& \ci{92.15}{91.06, 93.10}
& \ci{34.29}{33.94, 38.77} \\
& & Pix2Pix~\cite{isola2017image}
& \ci{92.21}{91.14, 93.14}
& \ci{34.07}{33.89, 38.66} \\
& & EKGAN~\cite{joo2023twelve}
& \ci{92.16}{91.07, 93.10}
& \cibest{34.60}{34.17, 38.98} \\
& & ECGRecover~\cite{lence2025ecgrecover}
& \ci{92.22}{91.16, 93.17}
& \ci{34.08}{33.96, 38.57} \\
& & ImputeECG
& \cibest{92.39}{91.32, 93.35}
& \ci{34.10}{34.04, 38.60} \\

\bottomrule
\end{tabular}
\end{table*}

\begin{table*}[t]
\centering
\caption{
\textbf{External validation of ECG waveform reconstruction on CPSC2018.}
Metrics are reported in the order of MAE, FD, SSIM, PSNR, MDD, ACD, MSE, and FID.
Best values within each ECG setting are shown in bold.
}
\label{tab:cpsc2018_reconstruction_8metrics}
\footnotesize
\setlength{\tabcolsep}{6pt}
\renewcommand{\arraystretch}{1.12}
\begin{tabular}{@{}llcccccccc@{}}
\toprule
\textbf{ECG setting} & \textbf{Methods}
& \textbf{MAE$\downarrow$} & \textbf{FD$\downarrow$} & \textbf{SSIM$\uparrow$} & \textbf{PSNR$\uparrow$}
& \textbf{MDD$\downarrow$} & \textbf{ACD$\downarrow$} & \textbf{MSE$\downarrow$} & \textbf{FID$\downarrow$} \\
\midrule

\multirow{5}{*}{4$\times$3 ECG}
& CycleGAN~\cite{zhu2017unpaired}
& 0.2757 & 0.3064 & 0.0926 & 11.0577 & 0.0052 & 0.1444 & 0.2115 & 0.0939 \\
& Pix2Pix~\cite{isola2017image}
& 0.1126 & 0.0369 & 0.2642 & 18.1654 & 0.0007 & 0.1629 & 0.0995 & 0.0014 \\
& EKGAN~\cite{joo2023twelve}
& 0.0974 & 0.0385 & 0.2491 & 19.5113 & 0.0006 & 0.1455 & 0.0680 & 0.0015 \\
& ECGRecover~\cite{lence2025ecgrecover}
& 0.1552 & 0.1556 & 0.2067 & 15.0701 & 0.0024 & 0.1028 & 0.1208 & 0.0242 \\
& ImputeECG
& \textbf{0.0490} & \textbf{0.0345} & \textbf{0.4520} & \textbf{27.1304}
& \textbf{0.0005} & \textbf{0.0459} & \textbf{0.0236} & \textbf{0.0012} \\
\midrule

\multirow{5}{*}{6$\times$2 ECG}
& CycleGAN~\cite{zhu2017unpaired}
& 0.2885 & 0.3421 & 0.0574 & 10.7184 & 0.0050 & 0.1164 & 0.2354 & 0.1171 \\
& Pix2Pix~\cite{isola2017image}
& 0.1286 & 0.0880 & 0.2476 & 17.2536 & 0.0014 & 0.1661 & 0.1257 & 0.0077 \\
& EKGAN~\cite{joo2023twelve}
& 0.1087 & \textbf{0.0338} & 0.2436 & 18.5958 & 0.0007 & 0.1498 & 0.0842 & \textbf{0.0011} \\
& ECGRecover~\cite{lence2025ecgrecover}
& 0.1594 & 0.1688 & 0.1613 & 15.0077 & 0.0025 & 0.1099 & 0.1374 & 0.0285 \\
& ImputeECG
& \textbf{0.0525} & 0.0361 & \textbf{0.3492} & \textbf{26.5731}
& \textbf{0.0005} & \textbf{0.0527} & \textbf{0.0240} & 0.0013 \\
\midrule

\multirow{5}{*}{12$\times$1 ECG}
& CycleGAN~\cite{zhu2017unpaired}
& 0.2208 & 0.3393 & 0.2222 & 11.8783 & 0.0018 & 0.0847 & 0.1860 & 0.1151 \\
& Pix2Pix~\cite{isola2017image}
& 0.0754 & 0.0181 & 0.5081 & 22.2440 & \textbf{0.0001} & 0.1105 & 0.0592 & 0.0003 \\
& EKGAN~\cite{joo2023twelve}
& 0.0780 & \textbf{0.0138} & 0.5137 & 22.0321 & 0.0014 & 0.1114 & 0.0513 & \textbf{0.0002} \\
& ECGRecover~\cite{lence2025ecgrecover}
& 0.0867 & 0.0315 & 0.4968 & 20.2845 & 0.0017 & 0.0810 & 0.0370 & 0.0010 \\
& ImputeECG
& \textbf{0.0363} & 0.0230 & \textbf{0.6098} & \textbf{30.6032}
& 0.0008 & \textbf{0.0317} & \textbf{0.0172} & 0.0005 \\
\bottomrule
\end{tabular}
\end{table*}

\begin{table*}[t]
\centering
\caption{
\textbf{External downstream diagnostic performance on CPSC2018.}
AUROC and AUPRC are reported as percentages with 95\% bootstrap percentile confidence intervals.
Complete and masked ECGs are included as reference baselines.
Bold values indicate the best reconstruction-based result within each incomplete ECG setting.
}
\label{tab:cpsc2018_downstream_ci}
\footnotesize
\setlength{\tabcolsep}{5pt}
\renewcommand{\arraystretch}{1.10}

\begin{tabular}{@{}lllcc@{}}
\toprule
\textbf{ECG setting} &
\textbf{Test input} &
\textbf{Method} &
\textbf{AUROC (\%, 95\% CI)} &
\textbf{AUPRC (\%, 95\% CI)} \\
\midrule

Complete ECG
& Ground truth
& --
& \ci{95.98}{95.21, 96.72}
& \ci{81.86}{79.64, 84.77} \\

\midrule
\multirow{6}{*}{\(4 \times 3\) ECG}
& Masked input
& No imputation
& \ci{89.02}{87.78, 90.17}
& \ci{63.98}{61.71, 66.63} \\
& \multirow{5}{*}{Reconstructed}
& CycleGAN~\cite{zhu2017unpaired}
& \ci{84.50}{83.03, 85.84}
& \ci{53.42}{51.44, 56.49} \\
& & Pix2Pix~\cite{isola2017image}
& \ci{90.63}{89.53, 91.67}
& \ci{65.72}{63.16, 68.82} \\
& & EKGAN~\cite{joo2023twelve}
& \ci{89.96}{88.67, 91.14}
& \ci{67.64}{65.10, 70.74} \\
& & ECGRecover~\cite{lence2025ecgrecover}
& \ci{91.17}{90.01, 92.21}
& \ci{69.02}{66.57, 72.02} \\
& & ImputeECG
& \cibest{94.75}{93.89, 95.58}
& \cibest{78.83}{76.67, 81.51} \\

\midrule
\multirow{6}{*}{\(6 \times 2\) ECG}
& Masked input
& No imputation
& \ci{93.72}{92.70, 94.68}
& \ci{76.46}{74.01, 79.31} \\
& \multirow{5}{*}{Reconstructed}
& CycleGAN~\cite{zhu2017unpaired}
& \ci{87.19}{85.83, 88.39}
& \ci{57.55}{55.29, 60.72} \\
& & Pix2Pix~\cite{isola2017image}
& \ci{92.93}{91.96, 93.90}
& \ci{72.65}{70.22, 75.86} \\
& & EKGAN~\cite{joo2023twelve}
& \ci{92.92}{91.95, 93.89}
& \ci{74.31}{71.73, 77.24} \\
& & ECGRecover~\cite{lence2025ecgrecover}
& \ci{92.99}{91.92, 93.96}
& \ci{73.36}{70.98, 76.47} \\
& & ImputeECG
& \cibest{95.05}{94.20, 95.84}
& \cibest{79.80}{77.64, 82.53} \\

\midrule
\multirow{6}{*}{\(12 \times 1\) ECG}
& Masked input
& No imputation
& \ci{95.65}{94.84, 96.45}
& \ci{81.34}{79.10, 84.19} \\
& \multirow{5}{*}{Reconstructed}
& CycleGAN~\cite{zhu2017unpaired}
& \ci{94.85}{93.97, 95.67}
& \ci{78.72}{76.38, 81.59} \\
& & Pix2Pix~\cite{isola2017image}
& \ci{95.69}{94.95, 96.44}
& \ci{80.94}{78.69, 83.86} \\
& & EKGAN~\cite{joo2023twelve}
& \ci{95.56}{94.77, 96.33}
& \ci{80.91}{78.60, 83.78} \\
& & ECGRecover~\cite{lence2025ecgrecover}
& \ci{95.55}{94.72, 96.36}
& \ci{81.38}{79.17, 84.24} \\
& & ImputeECG
& \cibest{95.89}{95.16, 96.62}
& \cibest{81.86}{79.67, 84.76} \\
\bottomrule
\end{tabular}
\end{table*}

\subsection*{ImputeECG Generalized to External CPSC2018 ECGs and Preserved Diagnostic Utility}

External testing on CPSC2018 showed that ImputeECG generalized beyond the PTB-XL development dataset and maintained strong waveform and diagnostic performance (Tables~\ref{tab:cpsc2018_reconstruction_8metrics} and~\ref{tab:cpsc2018_downstream_ci}). For waveform reconstruction, ImputeECG achieved the lowest MAE across the \(4 \times 3\), \(6 \times 2\), and \(12 \times 1\) incomplete ECG settings, with values of 0.0490, 0.0525, and 0.0363, corresponding to relative reductions of 49.7\%, 51.7\%, and 51.9\% compared with the strongest baseline in each setting. MSE showed a similar pattern, with relative reductions of 65.3\%, 71.5\%, and 53.5\%. ImputeECG also achieved the highest SSIM and PSNR and the lowest ACD across all three settings, indicating improved structural similarity, temporal consistency, and signal quality. Distributional metrics were more setting dependent, with ImputeECG achieving the best or near-best performance for MDD, FD, and FID in most settings.
% 在 CPSC2018 数据集上的外部测试表明，ImputeECG 能够推广到 PTB-XL 开发数据集之外的其他数据集，并保持了良好的波形和诊断性能（表 cpsc2018_reconstruction_8metrics 和 cpsc2018_downstream_ci）。对于波形重建，ImputeECG 在 4×3、6×2 和 12×1 三种不完整心电图设置下均实现了最低的平均绝对误差 (MAE)，其值分别为 0.0490、0.0525 和 0.0363，与各设置下最强的基线相比，相对降低了 49.7%、51.7% 和 51.9%。均方误差 (MSE) 也呈现类似的趋势，相对降低了 65.3%、71.5% 和 53.5%。在所有三种设置下，ImputeECG 均取得了最高的 SSIM 和 PSNR 值以及最低的 ACD 值，表明其在结构相似性、时间一致性和信号质量方面均有所提升。分布指标则更多地取决于设置，ImputeECG 在大多数设置下对 MDD、FD 和 FID 均取得了最佳或接近最佳的性能。

Downstream classification on CPSC2018 further showed that ImputeECG-completed ECGs preserved diagnostic information. Complete ground-truth ECGs achieved an AUROC of 95.98\% (95\% CI, 95.21--96.72) and an AUPRC of 81.86\% (95\% CI, 79.64--84.77). In the most incomplete \(4 \times 3\) setting, masked ECGs achieved an AUROC of 89.02\% (95\% CI, 87.78--90.17) and an AUPRC of 63.98\% (95\% CI, 61.71--66.63). After completion with ImputeECG, performance increased to 94.75\% AUROC (95\% CI, 93.89--95.58) and 78.83\% AUPRC (95\% CI, 76.67--81.51), corresponding to absolute gains of 5.73 and 14.85 percentage points over masked inputs. ImputeECG also exceeded the strongest reconstruction baseline in this setting by 3.58 percentage points in AUROC and 9.81 percentage points in AUPRC.
% 基于 CPSC2018 的下游分类进一步表明，使用 ImputeECG 完成的心电图保留了诊断信息。完整的真实心电图的 AUROC 为 95.98%（95% CI，95.21-96.72），AUPRC 为 81.86%（95% CI，79.64-84.77）。在最不完整的 (4 × 3) 设置下，掩蔽心电图的 AUROC 为 89.02%（95% CI，87.78-90.17），AUPRC 为 63.98%（95% CI，61.71-66.63）。使用 ImputeECG 后，性能提升至 94.75% AUROC（95% CI，93.89-95.58）和 78.83% AUPRC（95% CI，76.67-81.51），与掩蔽输入相比，绝对值分别提高了 5.73 和 14.85 个百分点。在此设置下，ImputeECG 的 AUROC 和 AUPRC 也分别比最佳重建基线高出 3.58 个百分点和 9.81 个百分点。

The diagnostic benefit remained consistent in the less severe missingness settings. In the \(6 \times 2\) setting, ImputeECG increased AUROC from 93.72\% to 95.05\% and AUPRC from 76.46\% to 79.80\%. In the \(12 \times 1\) setting, ImputeECG achieved 95.89\% AUROC and 81.86\% AUPRC, closely matching the complete-ECG reference. Across all three external classification settings, ImputeECG achieved the highest AUROC and AUPRC among reconstruction methods, supporting transferable reconstruction of clinically informative ECG signals under external testing.
% 在缺失值较少的情况下，诊断优势依然显著。在 6×2 的缺失值设置下，ImputeECG 将 AUROC 从 93.72% 提高到 95.05%，AUPRC 从 76.46% 提高到 79.80%。在 12×1 的缺失值设置下，ImputeECG 的 AUROC 和 AUPRC 分别达到 95.89% 和 81.86%，与完整心电图参考值高度吻合。在所有三种外部分类设置下，ImputeECG 在所有重建方法中均取得了最高的 AUROC 和 AUPRC，表明其在外部测试中能够有效地重建具有临床意义的心电信号。

\begin{figure*}[t]
    \centering
    \includegraphics[width=0.8\textwidth]{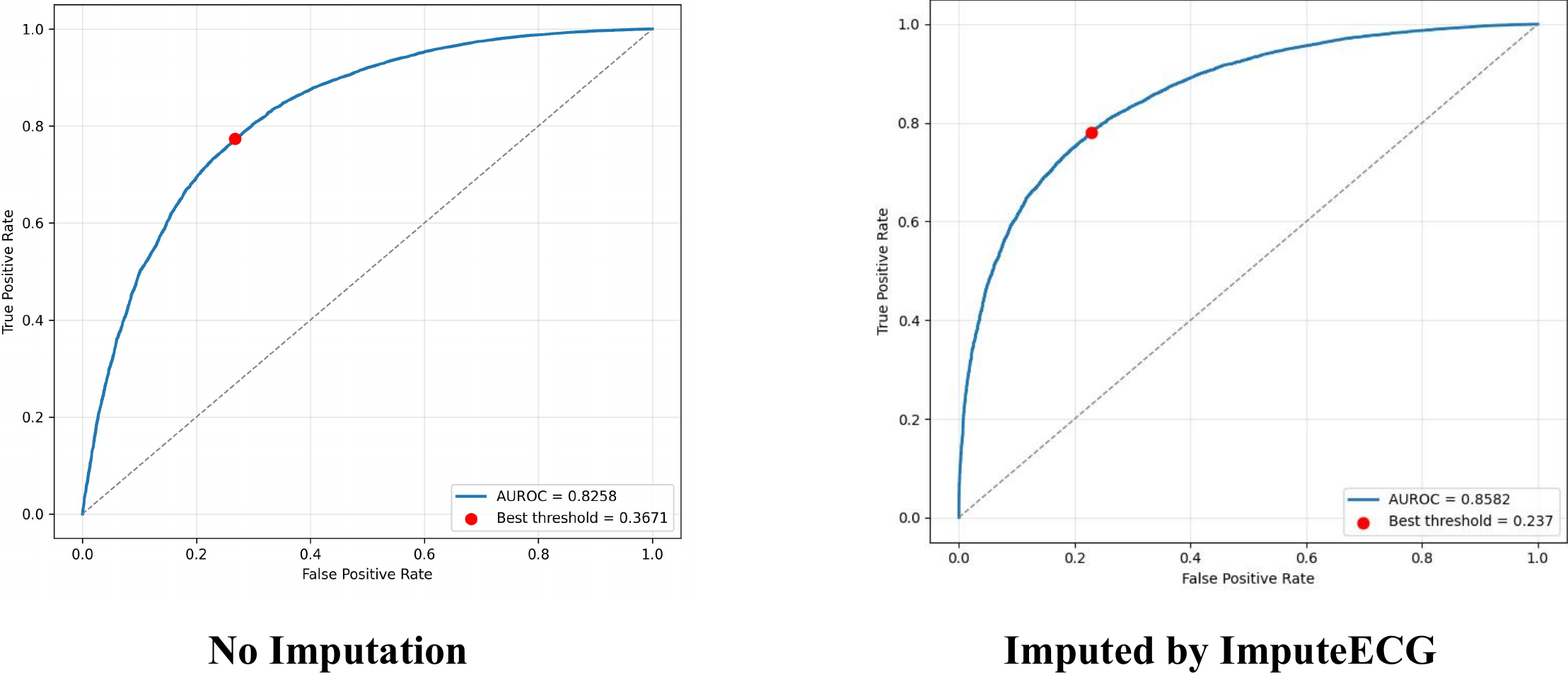}
    \caption{
    \textbf{Sex prediction performance in the independent clinical cohort.}
    The model performance was evaluated using completed ECGs and compared across input conditions.
    }
    \label{fig:kailuan_sex}
\end{figure*}

\begin{figure*}[t]
    \centering
    \includegraphics[width=0.8\textwidth]{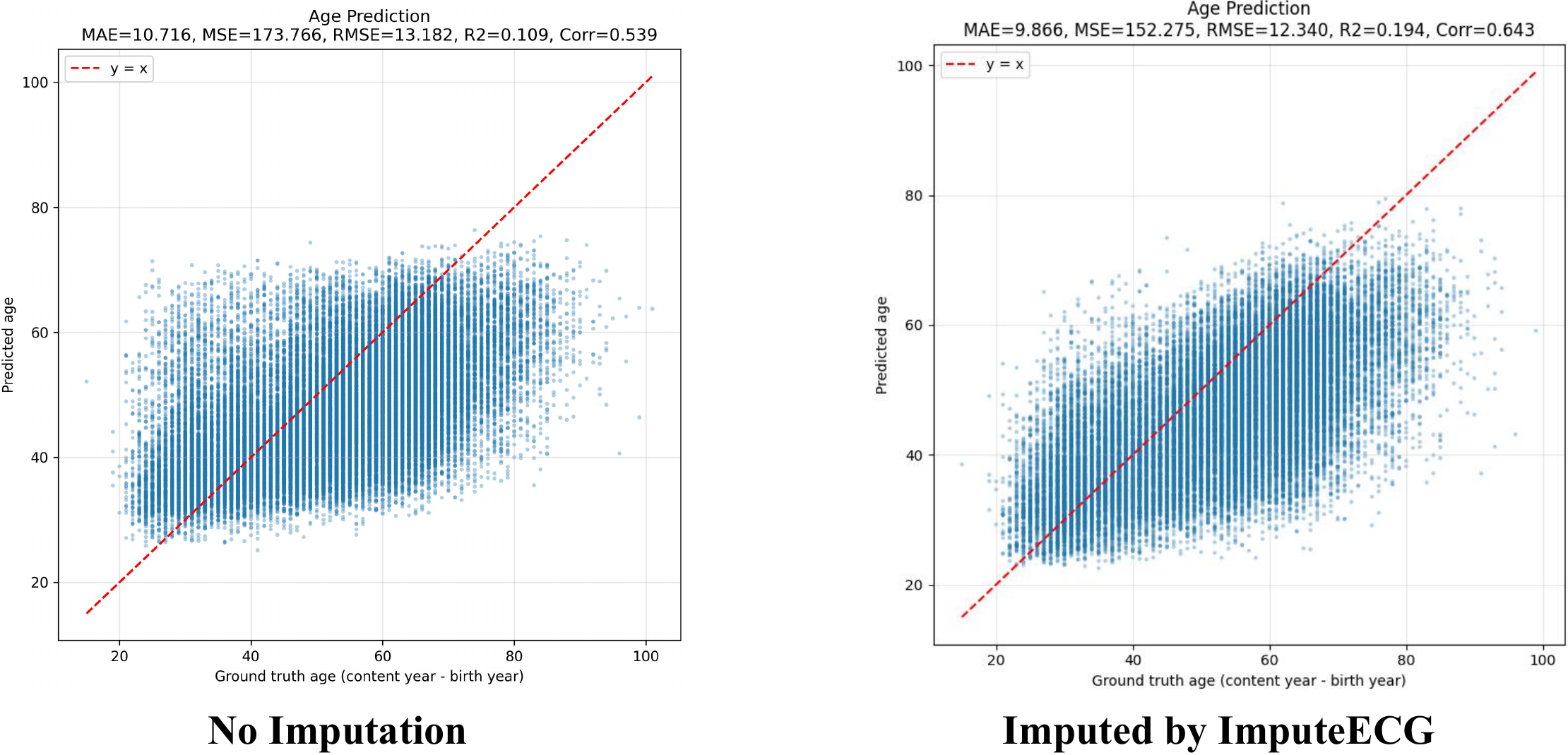}
    \caption{
    \textbf{Age prediction performance in the independent clinical cohort.}
    The model performance was evaluated using completed ECGs and compared across input conditions.
    }
    \label{fig:kailuan_age}
\end{figure*}

\subsection*{External Real-World Validation in the Kailuan Cohort}

We further evaluated ImputeECG in the independent Kailuan clinical cohort to assess its utility in a real-world ECG image digitization workflow. In this cohort, ECGs were originally available as image-based records. ECG images were digitized using an automated QRS-wave reconstruction-based ECG image-to-time-series algorithm\footnote{\url{https://github.com/PKUDigitalHealth/ecg-img2ts}}, and records that required additional processing were digitized using PaperECG\footnote{\url{https://github.com/Tereshchenkolab/paper-ecg}}~\cite{fortune2022digitizing}. The digitized Kailuan ECGs contained short-display formats, mainly \(4 \times 3\)-lead 2.5-second and \(6 \times 2\)-lead 5-second layouts, together with occasional unrecovered or corrupted waveform segments after image digitization. ImputeECG was therefore applied after digitization to generate complete 12-lead, 10-second ECG signals for downstream evaluation.
% 我们进一步在独立的开滦临床队列中评估了 ImputeECG，以检验其在真实世界心电图图像数字化工作流程中的实用性。在该队列中，心电图最初以图像记录的形式存在。心电图图像使用基于 QRS 波重建的自动心电图图像到时间序列算法进行数字化\footnote{\url{https://github.com/PKUDigitalHealth/ecg-img2ts}}，而需要额外处理的记录则使用 PaperECG\footnote{\url{https://github.com/Tereshchenkolab/paper-ecg}} 进行数字化。数字化后的开滦心电图包含短显示格式，主要为 4×3 导联 2.5 秒和 6×2 导联 5 秒的布局，图像数字化后偶尔会出现无法恢复或损坏的波形片段。因此，我们在数字化后应用 ImputeECG 生成完整的 12 导联 10 秒心电图信号，用于后续评估。

To assess whether ECG completion improved the usability of real-world digitized ECGs, we used a Net1D model~\cite{hong2020holmes} pretrained on the HEEDB dataset~\cite{koscova2026harvard} and applied it directly to the Kailuan cohort without cohort-specific fine-tuning. This zero-shot setting evaluated whether ImputeECG-completed ECGs could improve downstream prediction performance under external deployment conditions. Two clinically relevant prediction tasks were considered: sex prediction and age prediction. Performance was compared between digitized ECGs without imputation and ECGs completed by ImputeECG.
% 为了评估心电图补全是否能提高真实世界数字化心电图的可用性，我们使用了一个在 HEEDB 数据集上预训练的 Net1D 模型（hong2020holmes），并将其直接应用于开滦队列，而没有进行针对该队列的微调。这种零样本设置评估了 ImputeECG 补全后的心电图能否在外部部署条件下提高后续预测性能。我们考虑了两个具有临床意义的预测任务：性别预测和年龄预测。本研究比较了未经插补的数字化心电图和经 ImputeECG 补全的心电图的性能。

For sex prediction, ImputeECG improved AUROC from 0.8258 using non-imputed digitized ECGs to 0.8582 after ECG completion (Fig.~\ref{fig:kailuan_sex}). This corresponded to an absolute AUROC increase of 0.0324, indicating that reconstruction of missing waveform regions improved the discriminative information available to the pretrained classifier. For age prediction, ImputeECG also improved regression performance (Fig.~\ref{fig:kailuan_age}). Compared with non-imputed digitized ECGs, ImputeECG-completed ECGs reduced MAE from 10.716 to 9.866 years, MSE from 173.766 to 152.275, and RMSE from 13.182 to 12.340 years. The coefficient of determination increased from \(-0.109\) to 0.194, and the correlation between predicted and reference age increased from 0.539 to 0.643. These improvements indicate that ECG completion improved both absolute prediction accuracy and rank-level agreement in an independent real-world cohort.
% 对于性别预测，ImputeECG 将 AUROC 值从未经插补的数字化心电图的 0.8258 提高到补全后的 0.8582（图 ~\ref{fig:kailuan_sex}）。这相当于 AUROC 绝对值增加了 0.0324，表明缺失波形区域的重建提高了预训练分类器可用的判别信息。对于年龄预测，ImputeECG 也提高了回归性能（图 ~\ref{fig:kailuan_age}）。与未经插补的数字化心电图相比，经 ImputeECG 补全的心电图将 MAE 从 10.716 岁降低到 9.866 岁，MSE 从 173.766 降低到 152.275，RMSE 从 13.182 岁降低到 12.340 岁。决定系数从 -0.109 提高到 0.194，预测年龄与参考年龄的相关性从 0.539 提高到 0.643。这些改进表明，在独立的真实世界队列中，心电图补全提高了绝对预测准确率和等级一致性。

Together, the Kailuan validation demonstrates that ImputeECG improved downstream model performance after ECG image digitization, even when the downstream model was applied without cohort-specific fine-tuning. These findings support the practical value of ECG completion for converting incomplete digitized clinical ECG archives into more usable digital signals for AI-assisted cardiac assessment.
% 此外，开滦验证结果表明，即使在未对下游模型进行队列特异性微调的情况下，ImputeECG 也能在心电图图像数字化后提高下游模型的性能。这些发现支持了心电图补全在将部分数字化的临床心电图存档转换为更适用于人工智能辅助心脏评估的数字信号方面的实际价值。

\section*{Discussion}

In this study, we developed ImputeECG, a mask-conditioned one-dimensional Transformer autoencoder for completing incomplete 12-lead, 10-second ECGs while preserving all observed samples. Across clinically motivated missingness settings, ImputeECG improved waveform reconstruction on the internal PTB-XL test set and generalized to the external CPSC2018 dataset. The reconstructed ECGs showed lower point-wise errors, improved structural similarity and signal quality, and better preservation of morphology-related ECG features. These waveform-level gains were accompanied by improved downstream diagnostic performance, with completed ECGs restoring multi-label classification performance toward the complete-ECG reference in both internal and external evaluations. In an independent clinical cohort, ImputeECG also improved zero-shot sex and age prediction after image-based ECG digitization, supporting its practical value for real-world digital ECG workflows.
% 本研究开发了 ImputeECG，一种基于掩码条件的一维 Transformer 自编码器，用于在保留所有观测样本的同时，补全部分观测到的 12 导联 10 秒心电图。在临床相关的数据缺失场景下，ImputeECG 在内部 PTB-XL 测试集上提高了波形重建性能，并推广至外部 CPSC2018 数据集。重建后的心电图显示出更低的逐点误差、更高的结构相似性和信号质量，以及更好的形态学相关心电图特征保留。这些波形层面的提升伴随着下游诊断性能的提高，在内部和外部评估中，补全后的心电图均使多标签分类性能恢复到接近完整心电图的参考水平。在一个独立的临床队列中，ImputeECG 还提高了基于图像的心电图数字化后的零样本性别和年龄预测性能，验证了其在实际数字心电图工作流程中的应用价值。

This work addresses an important barrier in digital cardiovascular medicine. Many historical ECG records remain available only as paper printouts, scanned images, PDF reports, or short-display layouts. Although digitization algorithms can recover visible waveform traces, the resulting signals often remain incomplete because only short lead segments were displayed or because image-based recovery leaves corrupted or unrecovered waveform regions. This ECG incompleteness limits the reuse of ECG archives and reduces compatibility with ECG-AI models trained on complete digital waveforms. ImputeECG provides a completion step that converts incomplete ECG signals into standardized 12-lead, 10-second ECGs, thereby increasing the usability of heterogeneous clinical ECG archives for retrospective research and AI-enabled cardiovascular assessment.
% 这项工作旨在解决数字心血管医学领域的一个重要障碍。许多历史心电图记录目前仅以纸质打印件、扫描图像、PDF 报告或简短显示布局的形式存在。尽管数字化算法可以恢复可见的波形轨迹，但由于仅显示了较短的导联片段，或者基于图像的恢复方法会遗留损坏或无法恢复的波形区域，因此恢复后的信号通常只能部分可见。这种部分可见性限制了心电图档案的重用，并降低了与基于完整数字波形训练的心电图人工智能模型的兼容性。ImputeECG 提供了一个补充步骤，可以将部分可见的信号转换为标准化的 12 导联、10 秒心电图，从而提高异构临床心电图档案在回顾性研究和人工智能辅助心血管评估中的可用性。

The performance pattern across missingness settings suggests that ImputeECG effectively uses the physiological redundancy of 12-lead ECGs. The 12 leads capture related projections of the same cardiac electrical activity, while repeated cardiac cycles provide temporal context for reconstructing missing morphology. By explicitly conditioning on the missingness mask, the model can distinguish true low-amplitude ECG values from unavailable samples. The Transformer encoder--decoder further enables integration of long-range temporal context and cross-lead dependencies. This design was particularly beneficial in the \(4 \times 3\) and \(6 \times 2\) settings, where direct analysis of masked ECGs resulted in substantial information loss. The smaller gains in the \(12 \times 1\) setting are consistent with the fact that most lead-time information was already available and only local gaps required completion.
% 在不同缺失值设置下的性能模式表明，ImputeECG 有效地利用了 12 导联心电图的生理冗余性。12 个导联捕获了同一心脏电活动的相关投影，而重复的心动周期则为重建缺失的形态提供了时间上下文。通过显式地对缺失掩码进行条件化，该模型能够区分真实的低振幅心电图值和不可用的样本。Transformer 编码器-解码器进一步实现了长程时间上下文和跨导联依赖性的整合。这种设计在 4×3 和 6×2 设置下尤为有效，因为在这两种设置下，直接分析掩码心电图会导致大量信息丢失。在 12×1 设置下性能提升较小，这与大部分前置时间信息已经可用，仅需补充局部缺失值的事实相符。

The downstream analyses are central to the clinical interpretation of the study. Accurate waveform reconstruction alone provides incomplete evidence of diagnostic utility, because plausible-looking signals may still distort disease-relevant features. By applying fixed downstream classifiers to complete, masked, and reconstructed ECGs, we directly assessed whether completed signals retained clinically relevant information. On PTB-XL and CPSC2018, ImputeECG-completed ECGs consistently improved AUROC and AUPRC relative to masked inputs and achieved the strongest diagnostic performance among reconstruction methods. These results suggest that the model learned transferable temporal and inter-lead relationships rather than dataset-specific interpolation patterns.
% 下游分析是本研究临床解读的核心。仅凭精确的波形重建无法提供完整的诊断效用证据，因为看似合理的信号仍可能扭曲与疾病相关的特征。通过将固定的下游分类器应用于完整的、掩蔽的和重建的心电图，我们直接评估了完整信号是否保留了临床相关信息。在PTB-XL和CPSC2018数据集上，与掩蔽输入相比，使用ImputeECG完成的心电图始终提高了AUROC和AUPRC，并在所有重建方法中取得了最佳的诊断性能。这些结果表明，该模型学习的是可迁移的时间和导联间关系，而不是特定于数据集的插值模式。

The Kailuan cohort provides complementary real-world evidence. In this cohort, ECGs were digitized from image-based clinical records and included short-display layouts as well as unrecovered or corrupted waveform segments. ImputeECG improved sex prediction and age prediction using pretrained downstream models applied without cohort-specific fine-tuning. This zero-shot setting reflects a practical deployment scenario in which existing ECG-AI models are applied to newly digitized archives. The observed improvements indicate that ECG completion can make partially digitized ECG records more compatible with pretrained AI models.
% 开滦队列提供了补充性的真实世界证据。在该队列中，心电图数据来自基于图像的临床记录，包括短显示布局以及未恢复或损坏的波形片段。ImputeECG 通过使用预训练的下游模型，无需针对该队列进行微调，即可提高性别预测和年龄预测的准确性。这种零样本设置反映了将现有心电图人工智能模型应用于新数字化存档的实际部署场景。观察到的改进表明，心电图补全可以使部分数字化的心电图记录与预训练的人工智能模型更加兼容。

Several limitations should guide interpretation. First, the main reconstruction experiments used simulated missingness patterns generated from complete digital ECGs. These settings reflect common display formats and local waveform loss, but real-world ECG images may also contain grid artifacts, scanning distortion, compression artifacts, trace overlap, baseline drift, and low-resolution boundaries. Second, the Kailuan cohort provided downstream validation after digitization, but paired complete digital reference waveforms were unavailable for direct waveform-level evaluation. Third, reconstructed regions are model-inferred signals. Transient abnormalities, premature beats, ischemic changes, pacing artifacts, or noise events located entirely within missing regions may be difficult to recover from surrounding temporal and inter-lead context. Future work should incorporate uncertainty estimation, disease-specific safety analyses, and prospective evaluation across hospitals, ECG vendors, acquisition protocols, and digitization conditions.
% 以下几点局限性会影响结果的解读。首先，主要的重建实验使用了由完整数字心电图生成的模拟缺失模式。这些设置反映了常见的显示格式和局部波形丢失，但真实世界的心电图图像也可能包含网格伪影、扫描失真、压缩伪影、波形重叠、基线漂移和低分辨率边界。其次，开滦队列在数字化后提供了下游验证，但无法获得配对的完整数字参考波形进行直接的波形级评估。第三，重建区域是模型推断的信号。完全位于缺失区域内的瞬时异常、早搏、缺血性改变、起搏伪影或噪声事件可能难以从周围的时间和导联间上下文中恢复。未来的研究应纳入不确定性评估、疾病特异性安全性分析，并针对不同医院、心电图供应商、采集方案和数字化条件进行前瞻性评估。

Overall, ImputeECG demonstrates that mask-guided ECG completion can recover missing waveform information while preserving downstream diagnostic utility. By transforming incomplete or incompletely digitized ECG records into standardized 12-lead, 10-second signals, this approach may expand the usable scope of historical ECG archives and support more scalable AI-enabled cardiovascular assessment.
% 总体而言，ImputeECG 表明，基于掩模引导的心电图补全技术能够在保留后续诊断效用的同时，恢复缺失的波形信息。通过将不完整或部分数字化的心电图记录转换为标准化的 12 导联、10 秒信号，该方法有望扩展历史心电图档案的可用范围，并支持更具可扩展性的 AI 赋能心血管评估。

\newpage

\section*{Methods}
\subsection*{Datasets for Model Development and Evaluation}
Three ECG data sources were used to develop and evaluate ImputeECG: PTB-XL served as the model development cohort for training, validation, and internal testing; CPSC2018 was used for external assessment of waveform reconstruction and downstream diagnostic utility; and an independent Kailuan clinical cohort was used for real-world validation of ECG completion after image-based digitization.
% 我们使用了三个心电图数据源来开发和评估 ImputeECG：PTB-XL 作为模型开发队列，用于训练、验证和内部测试；CPSC2018 用于外部评估波形重建和下游诊断效用；独立的开滦临床队列用于在基于图像的数字化之后对心电图完成情况进行真实世界验证。

PTB-XL is a large publicly available 12-lead ECG dataset containing 21,799 10-second clinical ECG recordings from 18,869 subjects, with expert-annotated diagnostic labels~\cite{wagner2020ptb}. In this study, PTB-XL was used as the primary dataset for developing ImputeECG. Each complete 12-lead ECG served as the reference signal, and paired incomplete inputs were generated by applying predefined missingness patterns to the original waveform. All ECGs were represented as 12-lead, 10-second signals with 5000 time samples per lead. We used a subject-level data split, ensuring disjoint subjects across the training, validation, and internal test sets and preventing subject-level information leakage during model development and evaluation. After generating the incomplete-recording settings, the PTB-XL development set contained 52,254 paired examples for training, 6,549 for validation, and 6,594 for internal testing.
% PTB-XL 是一个大型的公开 12 导联心电图数据集，包含来自 18,869 名受试者的 21,799 条 10 秒临床心电图记录，并带有专家标注的诊断标签~\cite{wagner2020ptb}。在本研究中，PTB-XL 被用作开发 ImputeECG 的主要数据集。每条完整的 12 导联心电图均作为参考信号，通过将预定义的缺失模式应用于原始波形来生成成对的不完整输入。所有心电图均表示为 12 导联、10 秒的信号，每个导联包含 5000 个时间样本。我们采用受试者级别的数据划分，确保训练集、验证集和内部测试集中的受试者互不相交，从而防止在模型开发和评估过程中出现受试者级别的信息泄露。生成不完整记录设置后，PTB-XL 开发集包含 52,254 个用于训练的配对样本、6,549 个用于验证的配对样本和 6,594 个用于内部测试的配对样本。

CPSC2018 was used to assess external generalization~\cite{liu2018open}. The dataset contains 6,877 12-lead ECG recordings from 6,877 subjects. For waveform reconstruction evaluation, all 6,877 CPSC2018 ECGs were used as an external test set. The reconstruction model trained on PTB-XL was applied to CPSC2018 with fixed model parameters, and no CPSC2018 data were used to train or tune ImputeECG. The same incomplete-recording settings were generated from complete CPSC2018 recordings, allowing direct comparison between reconstructed ECGs and the corresponding complete reference signals. For downstream diagnostic evaluation, CPSC2018 was divided into training, validation, and test sets using a 7:1:2 subject-level split for classifier development and assessment. Subjects were disjoint across splits, preventing subject-level information leakage during downstream evaluation.
% CPSC2018 数据集用于评估外部泛化能力~\cite{liu2018open}。该数据集包含来自 6,877 名受试者的 6,877 条 12 导联心电图记录。为了评估波形重建能力，所有 6,877 条 CPSC2018 心电图均被用作外部测试集。在 PTB-XL 数据集上训练的重建模型被应用于 CPSC2018 数据集，并采用固定的模型参数。训练或调整 ImputeECG 模型时未使用任何 CPSC2018 数据。从完整的 CPSC2018 记录中生成了相同的不完整记录设置，从而可以直接比较重建的心电图和相应的完整参考信号。为了进行后续的诊断评估，CPSC2018 数据集被划分为训练集、验证集和测试集，划分比例为 7:1:2（受试者级别），用于分类器的开发和评估。受试者在划分过程中互不相交，从而防止在后续评估过程中出现受试者级别的信息泄露。

The Kailuan cohort was used as an independent real-world validation dataset to evaluate whether ECG completion improved the usability of incomplete clinical ECG data in an external setting. The cohort included 43,633 ECG records collected from three health examination waves between 2018 and 2022. Kailuan data were reserved exclusively for external evaluation: ImputeECG was applied with fixed reconstruction model parameters, and model training, model selection, and hyperparameter tuning were performed without using Kailuan records. This design ensured separation between model development and real-world validation and avoided data leakage. Because the Kailuan data reflected a clinical ECG image digitization scenario, they were used to assess whether incomplete and ImputeECG-completed ECG-derived inputs improved downstream clinical prediction tasks. Ethics approval and data-use governance for the Kailuan cohort are described in the Ethics statement.
% 开滦队列被用作一个独立的真实世界验证数据集，以评估心电图补全是否能提高不完整临床心电图数据在外部环境中的可用性。该队列包含2018年至2022年间三次健康体检中收集的43,633份心电图记录。开滦队列数据专门用于外部评估：使用固定的重建模型参数应用ImputeECG，并在不使用开滦队列记录的情况下进行模型训练、模型选择和超参数调优。这种设计确保了模型开发和真实世界验证的分离，并避免了数据泄露。由于开滦队列数据反映了临床心电图图像数字化场景，因此被用于评估不完整和经ImputeECG补全的心电图数据是否能改善下游临床预测任务。开滦队列的伦理审批和数据使用管理在伦理声明中进行了描述。

\subsection*{Data Processing}
For PTB-XL, we used the high-resolution 500~Hz version of the 10-second 12-lead ECG recordings and adopted the official PTB-XL patient-wise stratified split used in the benchmarking protocol~\cite{strodthoff2020deep}, with folds~1--8 for training, fold~9 for validation, and fold~10 for testing. ECG waveforms were retained at their original amplitude scale, and no z-score standardisation was applied.
% 对于 PTB-XL，我们使用了高分辨率 500 Hz 版本的 10 秒 12 导联心电图记录，并采用了基准测试方案中官方的 PTB-XL 患者分层数据集划分方法~\cite{strodthoff2020deep}，其中第 1-8 折用于训练，第 9 折用于验证，第 10 折用于测试。心电图波形保留了其原始振幅尺度，未进行 z 分数标准化。

For CPSC2018, we followed the ECG preprocessing pipeline used in ECGFounder~\cite{li2025electrocardiogram}. The 12-lead ECG recordings were represented at 500~Hz and filtered using a 0.5~Hz high-pass filter, a second-order Butterworth low-pass filter with a 50~Hz cutoff, and a 50/60~Hz notch filter. The filtered recordings were converted into 10-second segments: recordings longer than 10~seconds were split into consecutive 10-second windows, whereas recordings shorter than 10~seconds were zero-padded. Each 10-second segment was normalized using its own mean and standard deviation before model input.
% 对于 CPSC2018，我们采用了 ECGFounder~\cite{li2025心电图} 中使用的心电图预处理流程。12 导联心电图记录以 500 Hz 的采样率进行采样，并依次使用 0.5 Hz 高通滤波器、截止频率为 50 Hz 的二阶巴特沃斯低通滤波器以及 50/60 Hz 陷波滤波器进行滤波。滤波后的记录被转换为 10 秒的片段：长度超过 10 秒的记录被分割成连续的 10 秒窗口，而长度小于 10 秒的记录则进行零填充。每个 10 秒片段在输入模型之前都使用其自身的均值和标准差进行归一化。

For the Kailuan clinical cohort, image-based ECGs were digitised into time-series signals using an automated QRS-wave reconstruction-based ECG image-to-time-series algorithm, with PaperECG used for records requiring additional digitisation processing. The digitised ECGs mainly came from short-display formats, including \(4 \times 3\)-lead 2.5-second and \(6 \times 2\)-lead 5-second layouts. For signal preprocessing, each lead was processed independently using a 60~Hz notch filter to remove power-line interference, followed by a 0.5--50~Hz band-pass filter to attenuate baseline drift, low-frequency motion artefacts, muscle artefacts, and high-frequency electronic noise. Each lead was then z-score normalised within each ECG record before model input.
% 对于开滦临床队列，我们​​使用基于QRS波重建的自动心电图图像到时间序列转换算法，将图像心电图数字化为时间序列信号；对于需要额外数字化处理的记录，则使用PaperECG软件。数字化心电图主要来自短显示格式，包括4×3导联2.5秒和6×2导联5秒的布局。信号预处理方面，我们首先使用60 Hz陷波滤波器独立处理每个导联，以去除工频干扰；然后使用0.5~50 Hz带通滤波器衰减基线漂移、低频运动伪影、肌肉伪影和高频电子噪声。最后，在模型输入之前，我们对每个心电图记录中的每个导联进行z分数归一化。

\begin{figure*}[t]
    \centering
    \includegraphics[width=\textwidth]{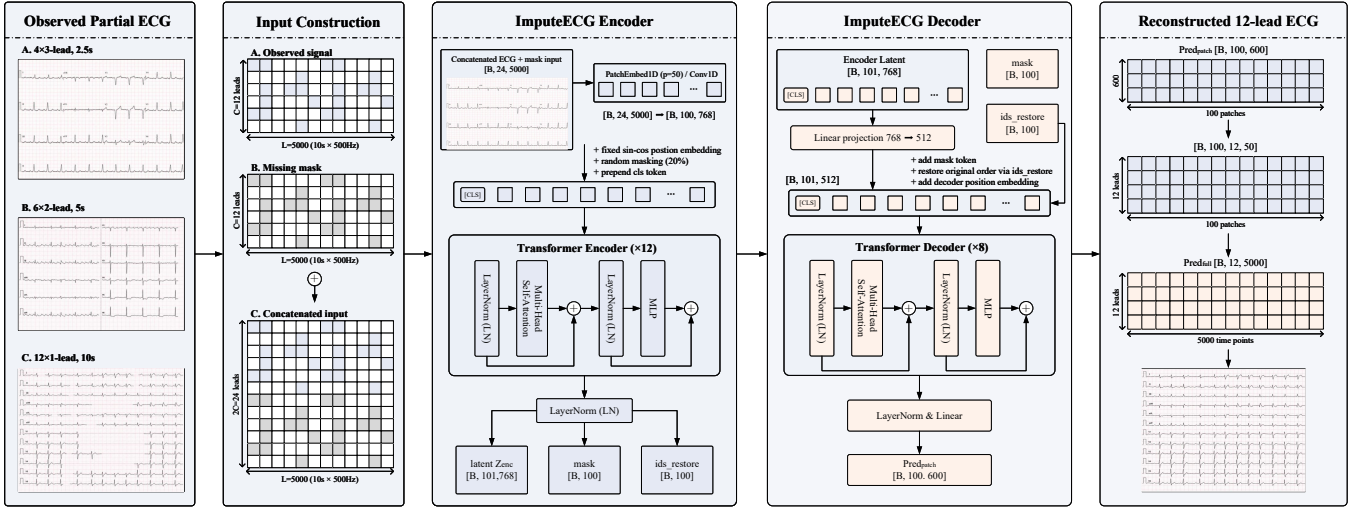}
    \caption{ \textbf{Architecture of ImputeECG.} Partially observed ECGs are combined with binary missingness masks and embedded as temporal patches. A one-dimensional Transformer encoder--decoder models temporal and inter-lead dependencies to reconstruct missing regions, generating a complete 12-lead, 10-second ECG while preserving observed samples. }
    \label{fig:framework}
\end{figure*}

\subsection*{ImputeECG Architecture}
ImputeECG was designed as a mask-conditioned one-dimensional Transformer autoencoder for reconstructing complete 12-lead, 10-second ECGs from incomplete recordings. For a mini-batch of ECGs, the complete reference waveform was denoted as \(X \in \mathbb{R}^{B \times C \times T}\), where \(B\) is the batch size, \(C=12\) is the number of ECG leads, and \(T=5000\) corresponds to a 10-second recording sampled at 500 Hz. The binary missingness mask was denoted as \(M \in \{0,1\}^{B \times C \times T}\), where \(M_{b,c,t}=1\) indicates that the sample at lead \(c\) and time point \(t\) in recording \(b\) is missing, and \(M_{b,c,t}=0\) indicates that the sample is observed. The availability mask was defined as \(A=\mathbf{1}-M\), where \(\mathbf{1}\) is an all-one tensor with the same shape as \(M\). The incomplete ECG input was obtained by retaining observed samples and zero-filling missing samples as \(X_{\mathrm{obs}}=A\odot X\), where \(\odot\) denotes element-wise multiplication.

The input to ImputeECG was formed by concatenating the zero-filled ECG waveform and the binary missingness mask along the lead dimension:
\begin{equation}
X_{\mathrm{in}}
=
\mathrm{Concat}(X_{\mathrm{obs}}, M;\mathrm{dim}=1)
\in
\mathbb{R}^{B \times 2C \times T}
=
\mathbb{R}^{B \times 24 \times 5000}.
\end{equation}
This signal--mask representation explicitly encodes sample availability, allowing the model to distinguish unavailable samples from physiologically meaningful low-amplitude ECG values.

The concatenated input was divided into non-overlapping temporal patches using a one-dimensional convolutional patch embedding layer. With patch size \(P=50\), each 10-second ECG was represented by \(N=T/P=100\) temporal patches, and \(D_e=768\) is the encoder embedding dimension. The patch embedding operation was formulated as:
\begin{equation}
Z_0
=
\mathrm{PatchEmbed}(X_{\mathrm{in}})
\in
\mathbb{R}^{B \times N \times D_e}.
\end{equation}

 A learnable class token \(z_{\mathrm{cls}}\) was prepended to the patch sequence, and fixed one-dimensional sine--cosine positional embeddings were added:
\begin{equation}
Z_0^{+}
=
[z_{\mathrm{cls}}; Z_0]
+
E_{\mathrm{pos}}^{e}
\in
\mathbb{R}^{B \times (N+1) \times D_e}
=
\mathbb{R}^{B \times 101 \times 768}.
\end{equation}

The encoder followed a Vision Transformer-style architecture adapted to one-dimensional ECG sequences. It consisted of \(L_e=12\) Transformer blocks, each containing multi-head self-attention, a feed-forward multilayer perceptron, residual connections, and layer normalization. For the \(l\)-th encoder block, the computation was defined as:
\begin{equation}
\tilde{Z}_{l}
=
Z_{l-1}
+
\mathrm{MSA}\!\left(\mathrm{LN}(Z_{l-1})\right),
Z_{l}
=
\tilde{Z}_{l}
+
\mathrm{MLP}\!\left(\mathrm{LN}(\tilde{Z}_{l})\right),
\qquad l=1,\ldots,L_e .
\end{equation}

The encoder used 12 attention heads and produced the latent representation \(Z_e=Z_{L_e}\in\mathbb{R}^{B \times 101 \times 768}\). This design enables the model to integrate long-range temporal dependencies across the full 10-second recording and cross-lead dependencies across the 12-lead ECG representation.

The decoder mapped the encoder output into a lower-dimensional reconstruction space. The encoder latent representation was first linearly projected from \(D_e=768\) to \(D_d=512\), followed by addition of fixed decoder positional embeddings:
\begin{equation}
H_0
=
Z_e W_d + b_d + E_{\mathrm{pos}}^{d}
\in
\mathbb{R}^{B \times 101 \times 512}.
\end{equation}

The decoder consisted of \(L_d=8\) Transformer blocks with 16 attention heads. For the \(k\)-th decoder block,

\begin{equation}
\tilde{H}_{k}
=
H_{k-1}
+
\mathrm{MSA}\!\left(\mathrm{LN}(H_{k-1})\right),
H_{k}
=
\tilde{H}_{k}
+
\mathrm{MLP}\!\left(\mathrm{LN}(\tilde{H}_{k})\right),
\qquad k=1,\ldots,L_d .
\end{equation}

After the decoder, the class token was removed, and the remaining \(N=100\) temporal tokens were passed to a linear prediction head:
\begin{equation}
Y_{\mathrm{patch}}
=
\mathrm{Linear}(H_{L_d}^{\mathrm{patch}})
\in
\mathbb{R}^{B \times N \times (C\cdot P)}
=
\mathbb{R}^{B \times 100 \times 600}.
\end{equation}

Each decoded token therefore predicted a 50-sample waveform segment for all 12 leads. The patch-level predictions were reshaped and concatenated along the temporal dimension to generate the reconstructed ECG:

\begin{equation}
\hat{X}
=
\mathrm{Unpatchify}(Y_{\mathrm{patch}})
\in
\mathbb{R}^{B \times C \times T}
=
\mathbb{R}^{B \times 12 \times 5000}.
\end{equation}

During training, the model learned to reconstruct the complete waveform \(X\) from the zero-filled input \(X_{\mathrm{obs}}\) and the corresponding missingness mask \(M\). The effective training mask combined predefined missing regions with additional randomly generated continuous temporal gaps:
\begin{equation}
M_{\mathrm{train}}
=
\mathrm{clip}
\left(
M_{\mathrm{pre}} + M_{\mathrm{gap}},
0,
1
\right),
\end{equation}
where \(M_{\mathrm{pre}}\) denotes the missing regions induced by the simulated incomplete ECG setting, and \(M_{\mathrm{gap}}\) denotes random continuous temporal gaps introduced as mask augmentation. This augmentation improves robustness to heterogeneous lead-time missingness patterns.

The reconstruction objective was an L1 loss computed only over missing samples:
\begin{equation}
\mathcal{L}_{\mathrm{rec}}
=
\frac{
\sum_{b=1}^{B}
\sum_{c=1}^{C}
\sum_{t=1}^{T}
M_{\mathrm{train},b,c,t}
\left|
\hat{X}_{b,c,t}
-
X_{b,c,t}
\right|
}{
\sum_{b=1}^{B}
\sum_{c=1}^{C}
\sum_{t=1}^{T}
M_{\mathrm{train},b,c,t}
+
\epsilon
},
\end{equation}
where \(\epsilon\) is a small constant for numerical stability. This missing-region objective directs optimization toward unobserved waveform segments and preserves the role of observed samples as conditioning information.

During inference, the model generated a 12-lead ECG prediction \(\hat{X}\), and the final completed ECG was constructed by retaining all originally observed samples exactly and filling the missing regions with the corresponding ImputeECG predictions:
\begin{equation}
X_{\mathrm{complete}}
=
A \odot X_{\mathrm{obs}}
+
M \odot \hat{X}.
\end{equation}

\subsection*{Implementation details}
For ImputeECG model training, all experiments were conducted on a single NVIDIA A100 GPU with 40~GB memory. ImputeECG was trained for 100 epochs using a batch size of 128. Model parameters were optimized with AdamW, using \(\beta_1=0.9\), \(\beta_2=0.95\), and a weight decay of 0.05. The base learning rate was \(1 \times 10^{-3}\), which was linearly scaled by batch size relative to a reference batch size of 256, giving an effective learning rate of \(5 \times 10^{-4}\). A learning-rate schedule with a 10-epoch warm-up phase was applied during training.

For downstream ECG classification, we used a 1D ResNet-style Net1D classifier~\cite{hong2020holmes} with 12 input channels, 64 base filters, stage widths of 64, 128, 256, and 512, two residual blocks per stage, a kernel size of 15, stride 2, and group width 16. For PTB-XL and CPSC2018, models were trained from scratch on 10-s, 12-lead ECG waveforms resampled to 500~Hz, corresponding to 5{,}000 samples per lead. The PTB-XL task included 71 diagnostic subclasses, whereas the CPSC2018 external validation task included 9 rhythm and conduction classes. Models were optimized for 50 epochs using AdamW with a learning rate of \(1 \times 10^{-3}\), a weight decay of \(1 \times 10^{-4}\), default beta parameters \(\beta_1=0.9\) and \(\beta_2=0.999\), a batch size of 64, and binary cross-entropy with logits loss.

To assess whether ECG completion improved the usability of real-world digitized ECGs, we further evaluated sex prediction and age prediction in the Kailuan clinical cohort. For these tasks, we used Net1D models pretrained on the HEEDB dataset~\cite{koscova2026harvard} and applied them directly to the digitized Kailuan ECGs in a zero-shot setting. This evaluation was designed to determine whether completed ECGs provided more usable physiological information for pretrained downstream models under an external, real-world deployment scenario.

\subsection*{Reconstruction Evaluation and Statistical Analysis}
Reconstruction performance was evaluated by comparing each completed ECG with the corresponding complete reference ECG. All reconstruction metrics were computed only over originally missing regions, as defined by the missingness mask, so that the assessment reflected the quality of imputed waveform segments. Metrics were calculated separately for each missingness scenario, dataset, and reconstruction method.

We used complementary signal-level metrics to quantify reconstruction fidelity, including marginal distribution difference (MDD), autocorrelation difference (ACD), mean squared error (MSE), mean absolute error (MAE), Fr\'echet distance (FD), Fr\'echet Inception Distance (FID), structural similarity index (SSIM), and peak signal-to-noise ratio (PSNR). Lower values indicate better reconstruction for MDD, ACD, MSE, MAE, FD, and FID, whereas higher values indicate better reconstruction for SSIM and PSNR. These metrics were selected to capture point-wise reconstruction error, distributional similarity, temporal correlation structure, waveform shape preservation, and signal-to-noise characteristics.

To assess clinically relevant morphology preservation, we further evaluated ECG waveform features derived from the reference and reconstructed signals. R-peak timing error, RR interval error, QRS duration error, and QT interval error were computed in milliseconds. Waveform-region reconstruction errors were summarized using MAE within P-wave, QRS-complex, and T-wave regions. The same ECG delineation pipeline was applied to reference and reconstructed ECGs across all methods.

\subsection*{Evaluation of Downstream Clinical Tasks}
To evaluate whether ECG completion preserved clinically relevant downstream information, we further assessed reconstructed ECGs using the downstream prediction pipelines described above. For PTB-XL and CPSC2018, the fixed Net1D multi-label classifiers were applied to complete reference ECGs, masked ECGs, and imputed ECGs generated by each reconstruction method. Classification performance was quantified using the macro-averaged area under the receiver operating characteristic curve, denoted as macro AUROC, and the macro-averaged area under the precision--recall curve, denoted as macro AUPRC, across target labels. For these two benchmark datasets, 95\% confidence intervals for macro AUROC and macro AUPRC were estimated using record-level bootstrap resampling with 1{,}000 replicates.

For the Kailuan clinical cohort, downstream utility was evaluated using pretrained Net1D models for sex prediction and age prediction. The same pretrained models were applied directly to digitized ECGs before and after completion. Sex prediction was evaluated using AUROC and AUPRC for the binary classification task. Age prediction was evaluated using mean absolute error (MAE), root mean squared error (RMSE), Pearson correlation coefficient, and coefficient of determination \(R^2\) between predicted and chronological age. The same input conditions, labels, model weights, and metric calculation procedures were used across reconstruction methods to ensure direct comparability.

\newpage

\section*{Extended Figures}
\begin{figure*}[h]
    \centering
    \includegraphics[width=\textwidth]{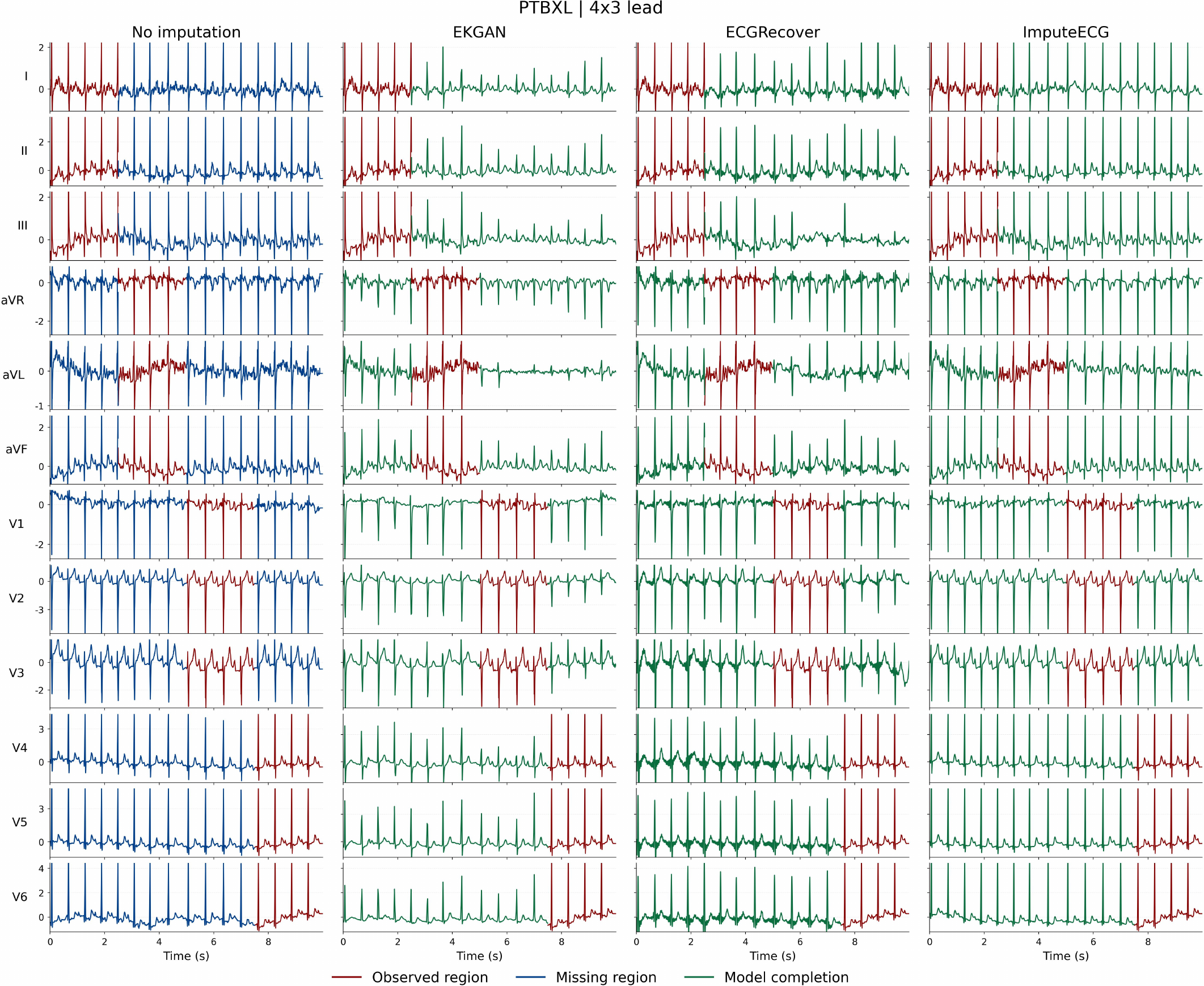}
\caption{
\textbf{Representative ECG completion example on PTB-XL.}
A 12-lead, 10-second ECG example from the \(4 \times 3\)-lead incomplete setting is shown for the masked input and reconstructions generated by EKGAN, ECGRecover, and ImputeECG.
Observed waveform regions are shown in red, missing regions in the uncompleted input are shown in blue, and model-completed regions are shown in green.
Compared with baseline methods, ImputeECG produced visually coherent waveform completions across leads and time.
}
\label{fig:4X3imputation}
\end{figure*}

\begin{figure*}[h]
    \centering
    \includegraphics[width=\textwidth]{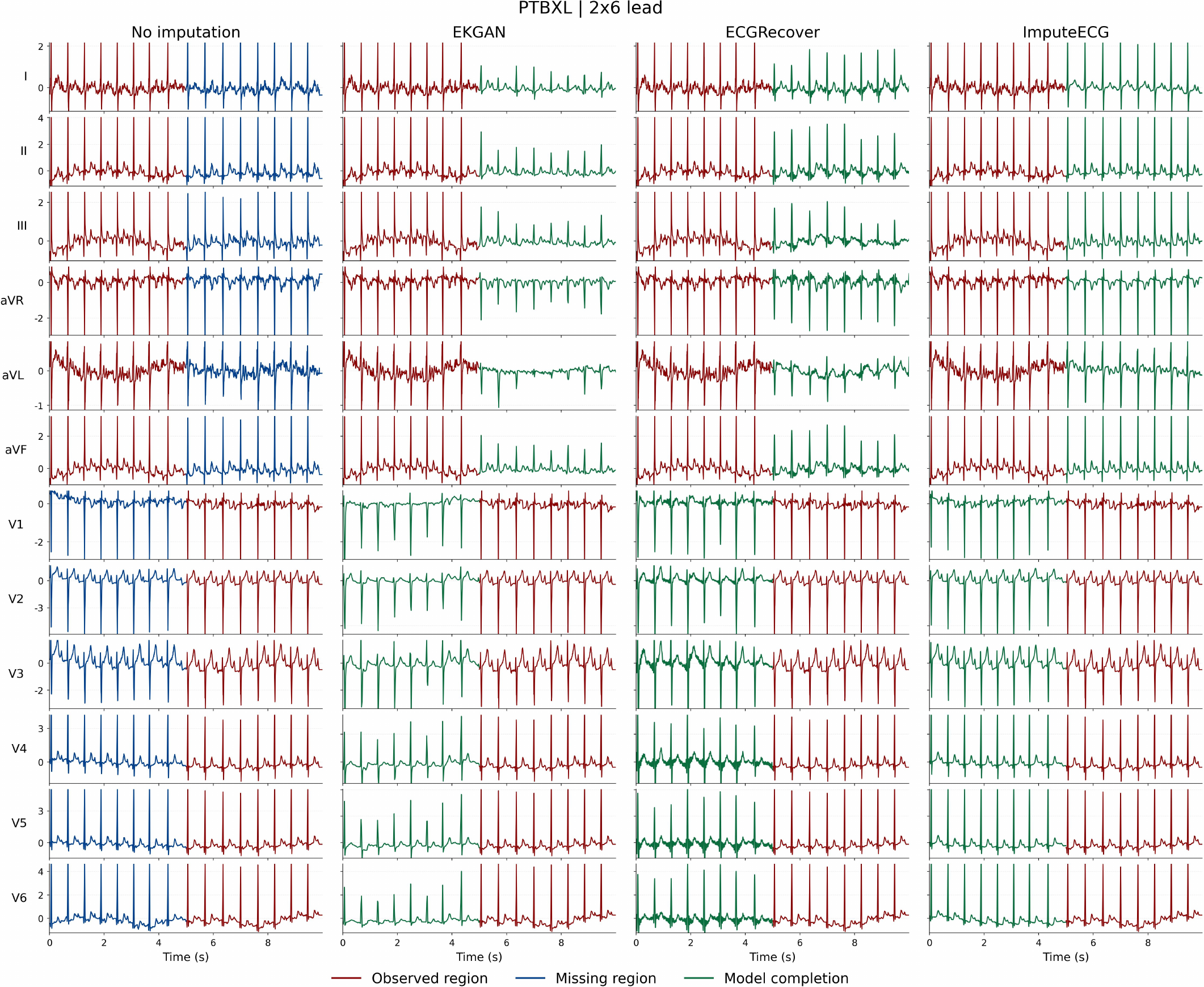}
    \caption{Representative ECG completion example on PTB-XL in the \(6 \times 2\)-lead setting.
}
    \label{fig:6X2imputation}
\end{figure*}

\begin{figure*}[h]
    \centering
    \includegraphics[width=\textwidth]{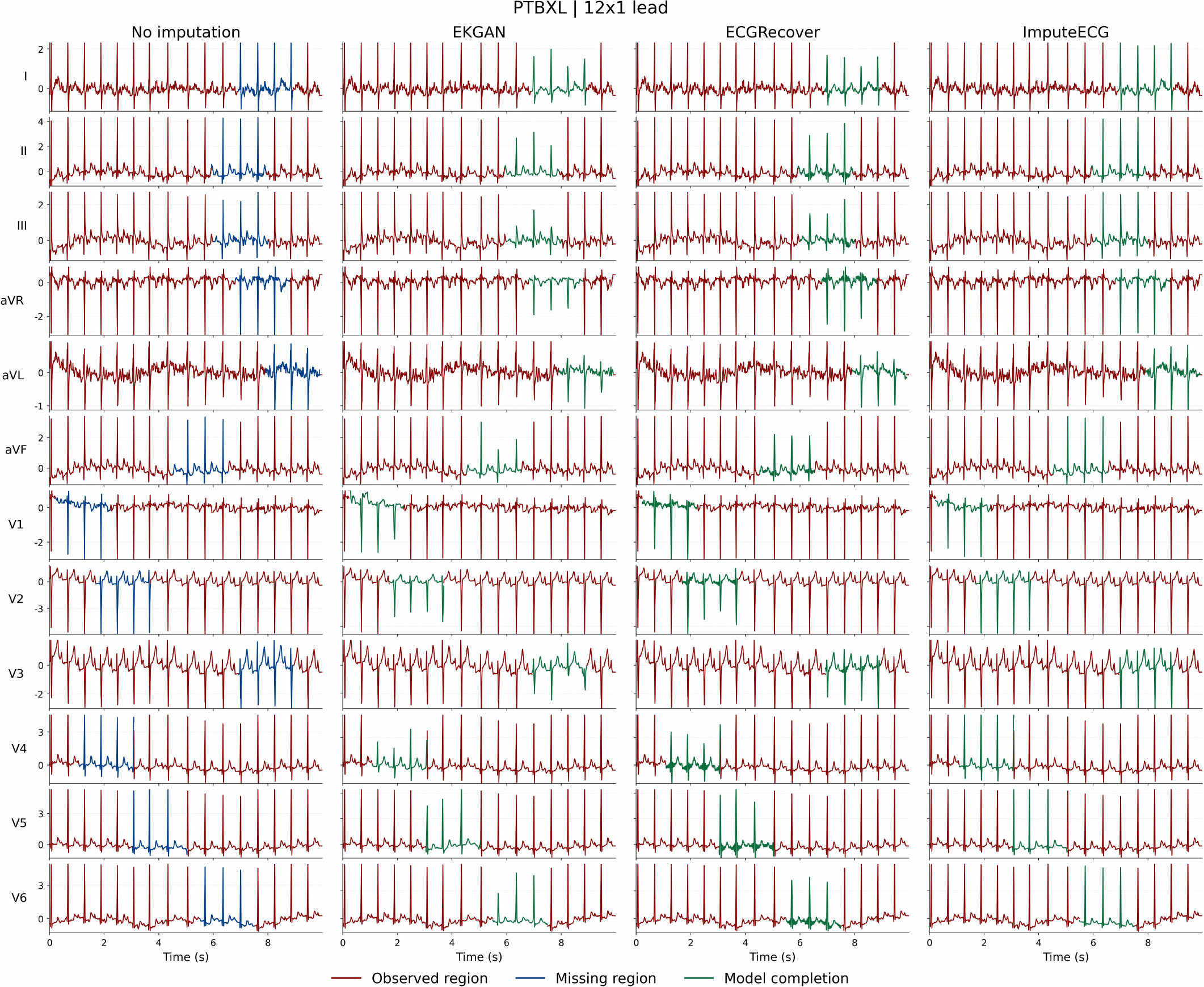}
    \caption{Representative ECG completion example on PTB-XL in the \(12 \times 1\)-lead setting.}
    \label{fig:12X1imputation}
\end{figure*}

\section*{Ethics Statement}
The Kailuan Study was approved by the Medical Ethics Committee of Kailuan General Hospital 
(approval number: [2006] Approval No. 5). 

\section*{Data Availability}
The public datasets used in this study are available from their respective repositories. PTB-XL is available from PhysioNet at \url{https://physionet.org/content/ptb-xl/1.0.3/}. CPSC2018 is available through the PhysioNet/Computing in Cardiology Challenge 2020 repository at \url{https://physionet.org/content/challenge-2020/1.0.2/training/cpsc_2018/}. The Kailuan clinical cohort is not publicly available because it contains patient-level clinical data subject to institutional data-use agreements, ethical approvals, and participant privacy restrictions. Access to the Kailuan data requires approval from the relevant study governance and ethics committees. 

\section*{Code Availability}
The source codes about this study and data analysis in this manuscript are provided at \url{https://github.com/PKUDigitalHealth/ImputeECG}.

\section*{Acknowledgments}
This work is supported by the National Natural Science Foundation of China (62102008), CCF-Tencent Rhino-Bird Open Research Fund (CCF-Tencent RAGR20250108), CCF-Zhipu Large Model Innovation Fund (CCF-Zhipu202414), PKU-OPPO Fund (BO202301, BO202503), Research Project of Peking University in the State Key Laboratory of Vascular Homeostasis and Remodeling (2025-SKLVHR-YCTS-02), Beijing Municipal Science and Technology Commission (Z251100000725008), Prevention and Control of Emerging and Major Infectious Diseases-National Science and Technology Major Project (2025ZD01906000, 2025ZD01906004), Capital’s Funds for Health Improvement and Research (CFH2026-1-4092), Beijing Natural Science Foundation (QY26080). 
%%%  Use this section to acknowledge contributions 
%%%  from non-authors and list funding sources, 
%%%  including grant numbers.

\section*{Author Contributions}
Xiaocheng Fang led the conceptualization of the study, designed the methodology, implemented the ImputeECG model, performed validation and formal analyses, and drafted the initial version of the manuscript. Haoyu Wang, Jieyi Cai, and Qinghao Zhao contributed to model development, experimental design, result analysis, and manuscript revision. Jun Li and Shanwei Zhang assisted with methodological refinement, downstream evaluation design, interpretation of results, and critical revision of the manuscript. Guangkun Nie, Yujie Xiao, Shun Huang, and Jiarui Jin contributed to data preprocessing, data curation, data cleaning, auditing, and investigation. Hongmin Liu, Guodong Wang, Shuohua Chen, Liming Lin, and Shouling Wu contributed to clinical data acquisition, cohort coordination, data curation, and clinical interpretation. Hongyan Li and Shenda Hong jointly supervised the project, provided resources, guided the study design, methodology development, and result interpretation, served as corresponding authors, and take full responsibility for the integrity of the work. All authors reviewed, revised, and approved the final manuscript.

\section*{Competing Interests}
The authors declare no competing interests.

\newpage

%%%  REFERENCES: As of 2023, all Cell Press journals 
%%%  use Numbered (AMA) style. We recommend placing 
%%%  your references in the included "references.bib" 
%%%  file.

\bibliography{reference}

\bigskip

%%%  In your References, please include only articles 
%%%  that are published (online publication and 
%%%  preprint servers are OK). Unpublished data, 
%%%  submitted and/or accepted manuscripts, abstracts, 
%%%  and personal communications should be cited within 
%%%  the text only ("unpublished data," "data not 
%%%  shown," "Alice Smith, personal communication") 
%%%  and not included in the references list. Personal 
%%%  communication should be documented by a letter 
%%%  of permission. Whenever possible, please make 
%%%  sure your .bib file has the complete author lists 
%%%  for each item (at minimum, the first 11 authors 
%%%  listed). 

\newpage
\begin{appendices}
\onecolumn

\end{appendices}

\end{document}